\documentclass[runningheads]{llncs}

\usepackage{eccv}

\usepackage{eccvabbrv}

\usepackage{graphicx}
\usepackage{booktabs}

\usepackage[accsupp]{axessibility}  

\usepackage{hyperref}

\usepackage{orcidlink}
\usepackage{times}
\usepackage{epsfig}
\usepackage{graphicx}
\usepackage{amsmath}
\usepackage{amssymb}
\usepackage{booktabs}
\usepackage{lineno}
\usepackage{tabu}
\usepackage{multirow}
\usepackage{svg}
\usepackage{mathrsfs}
\usepackage{CJK}
\usepackage{xcolor}
\usepackage{algorithm}
\usepackage{algpseudocode}
\usepackage{booktabs}
\usepackage{subcaption}

\begin{document}

\title{BIMM: Brain Inspired Masked Modeling for Video Representation Learning} 

\titlerunning{BIMM: Brain Inspired Masked Modeling for Video Representation Learning}

\author{Zhifan Wan\inst{1,2}\orcidlink{0000-1111-2222-3333} \and
Jie Zhang\inst{1}\orcidlink{1111-2222-3333-4444} \and
Changzhen Li\inst{1}\orcidlink{2222--3333-4444-5555}\and
Shiguang Shan\inst{1}\orcidlink{2222--3333-4444-5555}}

\authorrunning{Zhifan W, Jie Z.~Author et al.}

\institute{Institute of Computing Technology, Chinese Academy of Sciences, No.6 Kexueyuan South Road Zhongguancun, Haidian District, 100190, Beijing, China \and
University of Chinese Academy of Sciences, 19A Yuquan Rdn19A Yuquan Rd, Shijingshan District, 100049, Beijing, China
}

\maketitle

\begin{abstract}
The visual pathway of human brain includes two sub-pathways, \ie, the ventral pathway and the dorsal pathway, which focus on object identification and dynamic information modeling, respectively. 
Both pathways comprise multi-layer structures, with each layer responsible for processing different aspects of visual information. Inspired by visual information processing mechanism of the human brain, we propose the Brain Inspired Masked Modeling (BIMM) framework, aiming to learn comprehensive representations from videos. Specifically, our approach consists of ventral and dorsal branches, which learn image and video representations, respectively. Both branches employ the Vision Transformer (ViT) as their backbone and are trained using masked modeling method.
To achieve the goals of different visual cortices in the brain,
we segment the encoder of each branch into three intermediate blocks and reconstruct progressive prediction targets with light weight decoders. Furthermore, drawing inspiration from the information-sharing mechanism in the visual pathways, we propose a partial parameter sharing strategy between the branches during training. Extensive experiments demonstrate that BIMM achieves superior performance compared to the state-of-the-art methods. 
The code can be viewed in \textcolor{blue}{https://github.com/TonyAlbertWan/BIMM}.

\end{abstract}

\section{Introduction}
\label{sec:intro}

The Transformer, as introduced by Vaswani et al. \cite{vaswani2017transformer}, has emerged as a dominant force in the realm of natural language processing (NLP) \cite{kenton2019bert, brown2020language, radford2018improving}, exerting a substantial influence on the domain of computer vision (CV). 
The vision transformer (ViT) \cite{dosovitskiy2020vit} improves a variety of CV tasks including image classification \cite{touvron2021training, zhou2021deepvit}, object detection \cite{carion2020end, liu2021swin}, semantic segmentation \cite{xie2021segformer}, objective tracking \cite{chen2021transformer,cui2022mixformer}, and video recognition \cite{arnab2021vivit,bertasius2021space}. 
However, training effective ViTs typically requires large-scale annotated datasets, which is becoming increasingly expensive.
Fortunately, the Internet provides copious amounts of unlabeled data that can be directly utilized for unsupervised feature representation learning, thereby catalyzing the advancement of self-supervised learning methods.

Recently, following the masked language modeling (MLM) methods \cite{brown2020language,kenton2019bert} in NLP domain, masked image modeling (MIM) methods \cite{bao2021beit,he2022mae,xie2022simmim} achieve remarkable success in the field of image representation learning. MIM learns semantic representations by first masking some patches of the input image and then predicting the signals based on the unmasked parts, \eg, RGB pixels \cite{he2022mae,xie2022simmim}, discrete tokens \cite{bao2021beit}, or CLIP features \cite{ren2023deepmim}. 
This idea is soon adopted in the video representation learning domain, giving rise to the development of masked video modeling (MVM) methods, \eg, VideoMAE \cite{tong2022videomae} and MAE-ST \cite{feichtenhofer2022maest}. These methods, similar to MIM approaches, treat videos as 3D patches and predict the masked feature to learn video representation. 

\begin{figure}[!t]
  \centering
  \includegraphics[width=0.5\textwidth]{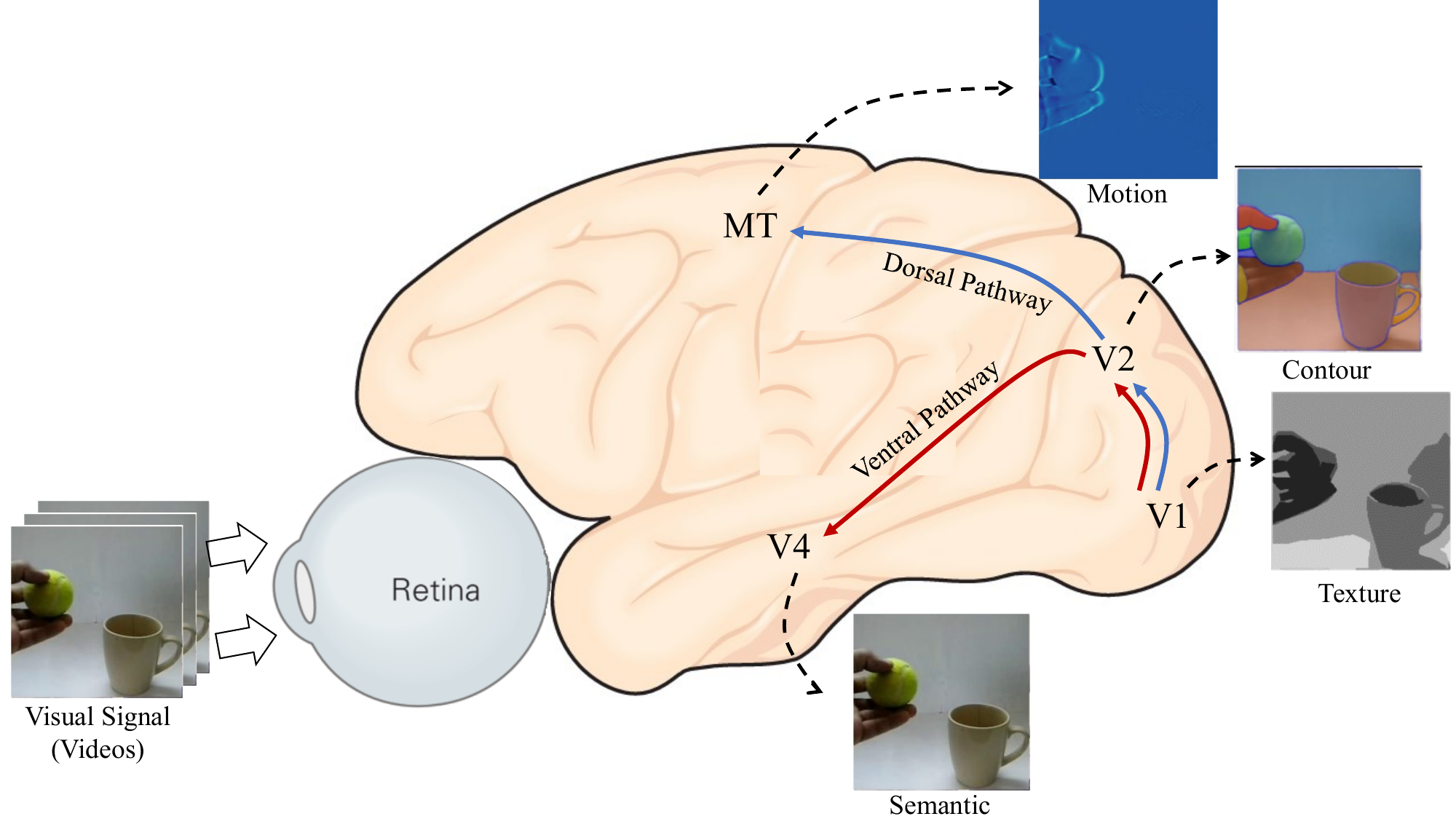}
  \caption{Visual pathways in the cerebral cortex. The four main areas in the visual pathway are denoted as  V1, V2, V4, and MT. Each area specialize in an aspect of visual information processing.}
  \label{fig:pathway}
  \vspace{-0.7cm}
\end{figure}

For human brain, the visual processing progress shares some similarities with masked modeling methods, both of which learn visual information from various vision signals.
However, the visual processing progress in brain is more complex.
The visual system of human brain comprises two main neural pathways: the ventral and dorsal pathways. 
As illustrated in Figure.\ref{fig:pathway}, these two pathways are represented as red and blue arrows, respectively.
The ventral pathway focuses on object identification, processing information about form and color. Meanwhile, the dorsal pathway is specialized in visually guided movement, featuring cells that are selective for the direction of movement . 
The four main areas in the visual pathway are labeled as V1, V2, V4, and MT, respectively \cite{kandel2000principles}.
The ventral and dorsal pathways share the first two areas, \ie, V1 and V2, which then relay information to the V4 and MT areas, respectively.
The primary visual area (V1) is sensitive to basic visual cues, \eg, light intensity and texture. 
The secondary visual area (V2) further processes the information received from V1, beginning to recognize object contours and orientations \cite{dehsorkh2023predicting, chen2021anatomical}.
The fourth visual area (V4) is instrumental in processing color and shape information and recognizing object characteristics, playing a critical role in our perception of the vivid world.  
The middle temporal area (MT) is primarily dedicated to processing visual motion \cite{schiller1993effects, li2019adaptive}. 
This dual-stream architecture enables the brain to process visual information efficiently and support dynamic visual behaviors.

\begin{figure*}[!t]
\vspace{-.3cm}
  \centering
  \makebox[\textwidth][c]{
  \includegraphics[width=1.05\textwidth]{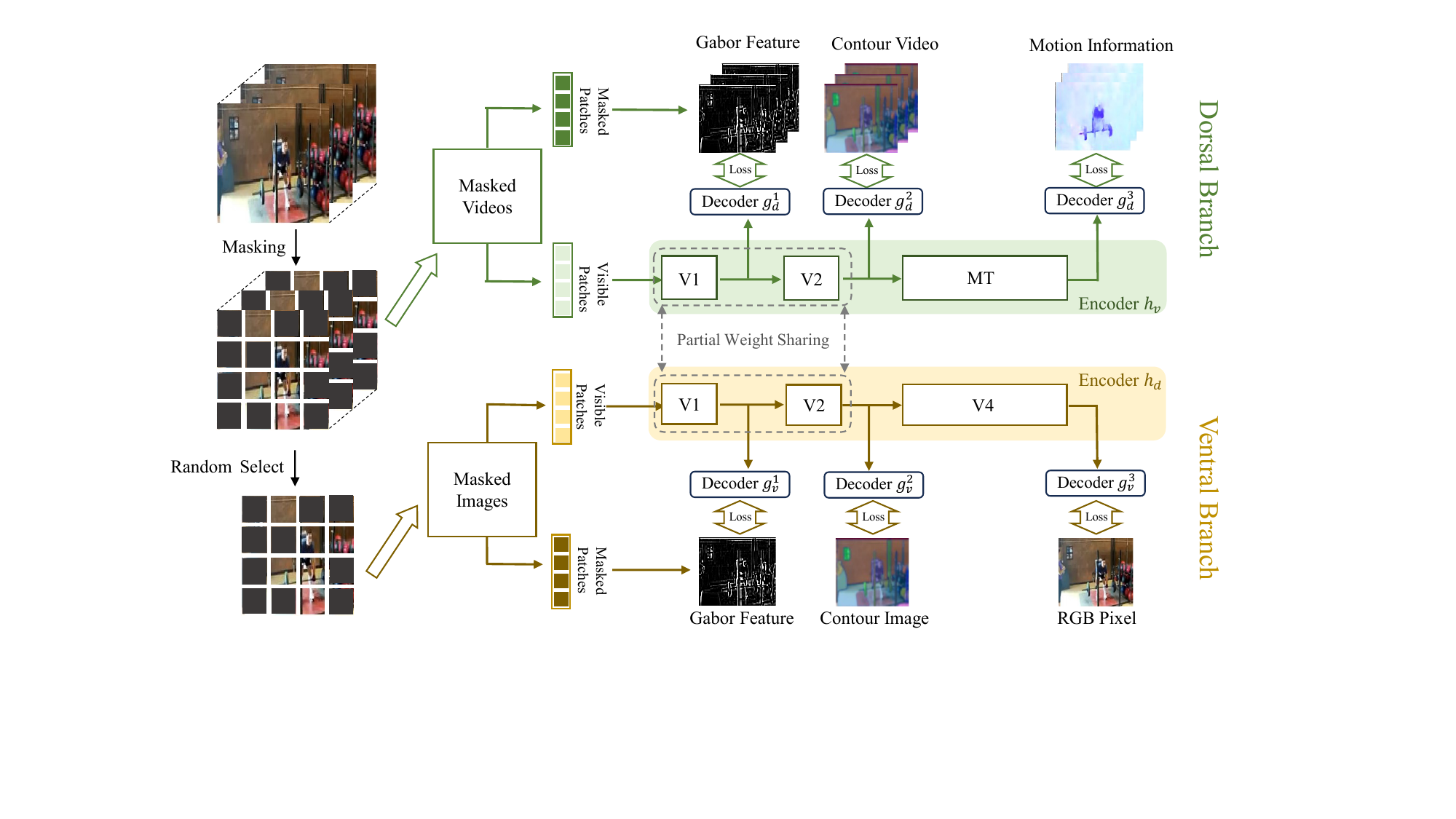}}
  \vspace{-2.2cm}
  \caption{
  An overview of BIMM. BIMM maintain a ventral branch and a dorsal branch, each employs masked modeling method. 
  Drawing inspiration from the visual pathways, ViT of each branch is divided into three intermediate blocks. 
  Each of these blocks is appended with a light-weight decoder. 
  Intermediate blocks (denoted as V1, V2, V4 and MT) are responsible for learning specific aspects of visual representation. 
  V1 is responsible for learning texture and predicts Gabor feature \cite{viola2001harr}.
  V2 specializes in contour detection, learns from contour images generated by SAM \cite{kirillov2023sam}.
  V4 is dedicated to color and object segmentation, with the prediction target being RGB pixels.
  MT is concerned with dynamic motion and predicts motion information \cite{yang2022motionmae}.
  During pretraining, BIMM applies a partial weight sharing strategy between the branches. 
  }
  \label{fig:pipline}
  \vspace{-0.7cm}
\end{figure*}

In light of this, we introduce the Brain Inspired Masked Modeling (BIMM), which aims to learn comprehensive video representations by simulating the visual information processing mechanism of the human brain. 
The detailed structure of BIMM is illustrated in Figure. \ref{fig:pipline}. 
Drawing inspiration from the dual-stream architecture in visual pathway, we employ one Vision Transformer (ViT) \cite{dosovitskiy2020vit} with MIM as our ventral branch, and another ViT based on MVM serves as the dorsal branch.
BIMM takes image-video pairs as input for the two branches, each of which follows an encoder-decoder structure \cite{he2022mae}.
Inspired by the specialized areas in visual pathways that process different aspects of visual information, we divide the ViT in each branch into three intermediate blocks. Each of these blocks is appended with a light-weight decoder. 
The feature extracted from each intermediate block is supervised by computing reconstruction losses, which correspond to the progressive prediction targets.
As it is widely used in image texture analysis, BIMM attaches Gabor feature \cite{gabor1946theory} to the first intermediate block to simulate the function of V1.
For the second intermediate block, we use the Segment Anything Model (SAM) \cite{kirillov2023sam} to extract contour of objects, forming a contour image. This mirrors the role of V2, which emphases shapes and orientations. 
In the third block, we utilize RGB pixels to learn color and shape information for basic object recognition. 
These three blocks together form our ventral branch, which is responsible for learning spatial knowledge.
The first two prediction targets of our dorsal branch are the same as those of the ventral branch, while the last block focuses on learning dynamic motion.
In more detail, the fourth block learns from reconstructing motion information \cite{yang2022motionmae}, defined as the pixel-level difference between two temporally nearby frames.
Considering two pathways share the information processing in the first two areas, we additionally design a partial parameter sharing strategy to facilitate information transfer between the ventral and dorsal branches.

For pretraining, the ventral branch learns spatial priors by training on ImageNet-1K dataset \cite{russakovsky2015imagenet} with Masked AutoEncoder (MAE) \cite{he2022mae} approach. 
Subsequently, the attention weight matrices of the dorsal branch are initialized using the pretrained ventral branch.
The two branches are then jointly trained on video data utilizing the partial weight sharing strategy, which shares the parameters of the first two intermediate blocks.
This design not only maintains static spatial representations acquired from the ventral branch but also learns dynamic motion information from the dorsal branch. 
For finetuning, the ViTs in BIMM are tuned on specific datasets for downstream tasks.
We select three popular video datasets for pretraining, \ie, Kinetics-400 (K400) \cite{kay2017kinetics}, Something-Something-v2 (SSv2) \cite{goyal2017ssv2}, and UCF101 \cite{ucf101}. BIMM achieves state-of-the-art performance on various evaluation datasets. Additionally, extensive ablation studies further demonstrate the effectiveness of our framework.

Overall, we summarize the contributions as follows:
1) Drawing inspiration from the visual information processing progress in the human brain, we introduce a novel BIMM method, which fully unleashes the power of different visual signals for self-supervised video representation learning.
2) The key designs of BIMM, \ie,the dual-branch structure, progressive prediction targets, and partial weight sharing strategy, enable the network to learn comprehensive video representations effectively.
3) We achieve state-of-the-art performance on various downstream tasks including video action recognition and action detection.

\section{Related Works}
\label{sec:relate}

\noindent\textbf{Self-supervised video representation learning.}
Self-supervised pretraining aims to encourage the model to learn transferable representations through a pretext task without labels. 
The early wave of self-supervised video representation learning methods rely on carefully designing time-related pretext tasks in three lines: predicting specific transformations (\eg, rotation angle \cite{jing2018self}, playback speed \cite{benaim2020speednet}, temporal order \cite{misra2016shuffle,lee2017unsupervised,xu2019self} and motion statistics \cite{wang2019self}), predicting future frame \cite{han2019video, han2020memory}, and instance discrimination \cite{qian2021spatiotemporal,wang2020self, chen2021rspnet}. 
Subsequently, researchers shift their focus to contrastive learning methods, which minimizes the similarity between different views of the same video while maximizing the dissimilarity between different videos \cite{chen2020simple,hadsell2006dimensionality,he2020momentum,xie2022simmim,feichtenhofer2021rho}. However, contrastive learning methods rely heavily on strong data augmentation and large batch size \cite{feichtenhofer2021rho}. 
Recently, inspired by the success of MLM \cite{brown2020language,kenton2019bert} and MIM \cite{bao2021beit,he2022mae,xie2022simmim}  methods in NLP and CV, there is growing interest in masking-based methods for self-supervised learning on videos. 

\noindent\textbf{Masked modeling.}
With the increasing adoption of Transformers \cite{vaswani2017transformer} in computer vision, masked modeling methods are emerged as a general self-supervised learning approach.
This trend is largely influenced by the success of BERT \cite{kenton2019bert} in a wide range of NLP downstream tasks.
Researchers soon discover that masked modeling is also effective in image representation learning.
Specifically, masked image modeling methods learn the representation by predicting the masked regions from visible parts. 
For instance, ViT \cite{dosovitskiy2020vit} predicts the mean colors of masked patches. BEiT \cite{bao2021beit} achieves enhanced performance with masked visual token prediction. Interestingly, MAE \cite{he2022mae} further demonstrates the effectiveness of image patch reconstruction. It improves pretraining efficiency by adopting a high mask ratio and encoding only unmasked patches. Alternatively, MaskFeat \cite{wei2022maskfeat} leverages HOG \cite{dalal2005hog} feature as the prediction target, resulting in robust visual representations.

Recently, masked modeling is applied for learning more challenging spatiotemporal representations from videos.
BEVT \cite{wang2022bevt} innovatively decouples masked modeling into two stages: spatial representation learning on images and temporal dynamics learning on both images and videos in a two-stage process.
VideoMAE \cite{tong2022videomae} and MAE-ST \cite{feichtenhofer2022maest} simply reconstruct masked spatiotemporal patches of each video and achieve strong performance on video recognition downstream tasks. MotionMAE \cite{yang2022motionmae} additionally predicts the corresponding motion knowledge by predicting masked motion information.
Furthermore, OmniMAE \cite{girdhar2023omnimae} contributes to unifying the image and video masked modeling method with one encoder. 
Overall, masked modeling proves to be an effective self-supervised video representation learning method, capable of learning representations from a wide range of modeling targets.
Consequently, we employ mask modeling method to simulate the information processing progress in the visual pathway, which requires to model different visual information.

\noindent\textbf{Brain-inspired methods in deep learning.} 
Many methods draw on knowledge from neuroscience to improve deep learning methods.
Spiking Neural Networks (SNNs)\cite{MAASS19971659} are probably the best-known brain-inspired deep learning method. SNN simulates the spiking signals between neurons, bridging the gap between neuroscience and deep learning. 
Furthermore, some methods apply neuroscience knowledge to simplify the complex representation learning process.
These include efforts in sound event recognition \cite{9093814}, video captioning \cite{yao2023brain}, and latent diffusion models \cite{Chen_2023Diffusion}.
Following this idea, we introduces a self-supervised masked modeling method combined with neuroscience knowledge, which aims to learn comprehensive visual representations.
A recent study \cite{Choi2024dual} proposes WhereCNN and WhatCNN to emulate the ventral and dorsal pathways in the human brain. 
It combines functional Magnetic Resonance Imaging (fMRI) images and attention maps from CNNs to demonstrate that the dual-pathway structure and the attention mechanisms of the visual pathways are similar to the attention mechanisms of neural networks.
Notably, although both BIMM and \cite{Choi2024dual} are inspired from the dual-branch structure in human brain visual processing, our goal is to use this knowledge for learning comprehensive visual representations from videos and images. In contrast, \cite{Choi2024dual} does not validate the performance of network in any downstream tasks.

\noindent\textbf{Intermediate blocks with multiple decoders.}  
Considering that different areas in the visual information processing focus on distinct visual information, the traditional structure of MVM methods \cite{wang2022bevt, wei2022maskfeat, tong2022videomae}, which use a single encoder and a single decoder, is insufficient to simulate the multi-layered structure in the human brain.
In fact, there is a long history of appending additional supervision signals to networks to facilitate the learning of multi-level visual information.
GoogleNet \cite{szegedy2015googlenet} and DSN \cite{wang2015dsn} introduce extra supervision objectives on intermediate layers to address training time and mitigate gradient vanishing / exploding \cite{hochreiter2001gradient} problems. 
Recent works \cite{ren2023deepmim, wang2023localmim} improve masked image modeling method by adding decoders to the intermediate blocks of the encoder, enabling multi-scale image feature learning. Such design reduces training time and enhances performance of pretrained model in downstream tasks.
In BIMM, we divide the encoder of each branch into three intermediate blocks and attach a decoder to each block for computing losses with various visual features.
The meticulously designed targets push each Transformer block to focus on different aspects of visual information, leading to better performance on various video understanding tasks.
Although BIMM and \cite{ren2023deepmim,wang2023localmim} all utilize multiple decoders for mask modeling methods, there are some significant differences:
1) BIMM is a comprehensive video representation learning work but DeepMIM \cite{ren2023deepmim} and LocalMIM \cite{wang2023localmim} foucs on masked image modeling. 
2) BIMM adopts progressive prediction targets in light of human visual information processing progress, while both DeepMIM and LocalMIM use a uniform reconstruction target, \eg, RGB pixels.

\section{Method}
In the initial stages of visual processing, information from the eye is soon relayed to the cortical areas after basic processing. 
Within these cortical areas, there are two distinct pathways: the ventral pathway and the dorsal pathway \cite{kandel2000principles}.
Drawing inspiration from this, BIMM comprises a ventral branch for image feature learning and a dorsal branch for video representation learning, as depicted in Figure. \ref{fig:pipline}. 
Specifically, each branch employs an encoder-multi-decoder architecture within the masked modeling framework for ViT pretraining. We utilize MAE \cite{he2022mae} and VideoMAE \cite{tong2022videomae} as baseline for ventral and dorsal branches, respectively.
Our method aims to effectively learn video representations by simulating the information processing progress of the human brain. 
In the subsequent sections, we will introduce the components of our framework.

\subsection{Architecture}
\noindent\textbf{Encoders.} 
The pathways in brain learn different signals of visual information with multiple cortex areas, each of which specialize for one aspect.
Drawing inspiration from this, we separate the encoder of each branch into three intermediate blocks.
We adopt Vision Transformer (ViT) \cite{dosovitskiy2020vit}, comprising 12 Transformer blocks, as encoder for each branch, denoted as $h_v$ and $h_d$, respectively. 
These encoders are segmented into three intermediate blocks for modeling two branches.
In the ventral branch, an input image $i\in \mathbf{R}^{H\times W\times C}$ is divided into regular non-overlapping patches.
These patches are randomly masked at a high ratio, producing a masked image. Visible patches $\hat{i}$ in masked images are then fed to the image encoder $h_v$ to generate multi-level image features from intermediate blocks.
For the dorsal branch, a video clip with $t$ frames, randomly sampled from the original video, undergoes temporal sampling to be compressed into $T$ frames with a temporal stride of $\tau$. 
The compressed video clip $v\in \mathbf{R}^{H\times W\times C\times T}$ is then adopted to the joint space-time cube embedding and random tube masking, creating the masked video clip. Visible patches $\hat{v}$ in masked videos are subsequently fed into the video encoder $h_d$ to produce multi-level video features from intermediate blocks. 
The features produced by the blocks are then input to decoders and calculate reconstruction loss with progressive prediction targets.

\noindent\textbf{Decoders.} 
To enable the intermediate layers to learn visual representations from the corresponding prediction targets, the features generated by each intermediate blocks should pass through a lightweight decoder.
In BIMM, we attach three decoders to the 12 Transformer blocks in each ViT of two branches.
Specifically, we incorporate three decoders at the $2nd$, $4th$, and $12th$ Transformer blocks, referred to as $g_v^1$, $g_v^2$, $g_v^3$ for ventral branch, and $g_d^1$, $g_d^2$, $g_d^3$ for dorsal branch, respectively.
For predicting various vision signals, each decoder consists of two parts: Transformer blocks for reasoning and multi-layer perceptron (MLP) for prediction. 
However, differing from other intermediate blocks, V1 predicts the Gabor feature, hence $g_v^1$ and $g_d^1$ only contain few linear layers.
Notably, the MLPs in these decoders are specifically customized to generate image and video patches.

\subsection{Progressive Prediction Targets}
In neuroscience, the four main areas in the cortical area, labeled as V1, V2, V4, and MT, each specialize in a distinct aspect of visual information processing.
The first area of the cortical areas is the early visual areas, which comprises two visual cortex areas called the primary visual area (V1) and the secondary visual area (V2).  
Both pathways utilize these early visual areas and channel visual information to distinct regions: the ventral pathway to the fourth visual area (V4) and the dorsal pathway to the middle temporal area (MT) \cite{kandel2000principles}. 
To simulate the information processing mechanism of these visual pathways, we develop BIMM with multiple decoders tailored to various visual signals, called progressive prediction targets.
This section will provide a detailed exploration of each prediction target.

\noindent\textbf{V1} is adept at processing local texture representations. 
Visual information received from eyes is firstly processed by V1, in which neurons selectively respond to depth and variations in light intensity \cite{dehsorkh2023predicting}. Therefore, V1 is crucial for interpreting material and texture during object recognition. 
Gabor features \cite{gabor1946theory}, generated by applying images to Gabor filters, play a crucial role in image texture analysis.
These features are adept at capturing local spatial frequencies, orientations, and scales within images, making them highly effective for texture analysis.
Therefore, the prediction target of the first intermediate blocks in both branches is the Gabor feature generated from the masked patches.

\noindent\textbf{V2} 
is another crucial component in early visual areas.
Neurons in V2 are sensitive to global disparity cues and respond to illusory contours created by adjacent line elements. 
Stating it alternatively, V2 discerns foreground-background relationships through the delineation of object contours \cite{schiller1993effects}.
We use the Segment Anything Model \cite{kirillov2023sam} to precisely segment the semantic contours of objects in an image, then concatenate all semantic masks together to form a "contour image". \footnote{We use SAM to effectively and accurately recognize object contours, which can be replaced by other edge detectors or digital image processing methods.}
These contour images are then patchfied and serve as the prediction target for the second intermediate blocks.
Notably, in the dorsal branch, edge images are produced for each frame of the video and subsequently amalgamated to create a "contour video".

\noindent\textbf{V4}
in ventral pathway integrates information about color and object shape, plays a key role in object recognition \cite{schiller1993effects}. 
The original image is composed of RGB pixels, containing color and object information. 
Following the practice in \cite{he2022mae}, we calculate the reconstruction loss between the outputs from the third intermediate blocks and their corresponding masked image patches. 

\noindent\textbf{MT} 
is responsible for the integration of the local motion signals.
Neurons in MT are selective for a particular direction of movement of an overall pattern \cite{li2019adaptive}, which contributes in the action recognition.
For the last decoder ($g_d^3$) of dorsal branch, we set the motion information \cite{yang2022motionmae} as the prediction target. 
The motion information is derived by calculating the L1 difference between pixel values of two temporally adjacent frames, effectively capturing the short-term temporal dynamics.
Such a simple target vividly showcases the overall motion pattern in the video.

\subsection{Training Strategy}
\noindent\textbf{Training objectives.} 
Most existing MVM models predict the supervision target $p$ based on the final output feature of the encoder $h$ and minimize a global reconstruction loss, \eg, MSE loss:
\begin{equation}
\begin{split}
  L_{MVM} = 
  ||g(h(\hat{v})) - p||_2^2 
  \end{split}
\end{equation}
where $g$ denotes the default decoder and $p$ denotes the prediction target.

BIMM contains two branches with multiple prediction targets. Therefore, the training loss of ventral branch and dorsal branch are denoted as $L_V$ and $L_D$:
\begin{equation}
\begin{split}
  L_V &= 
  ||g_v^1(h_v^1(\hat{i})) - p^{Gabor}_v||_2^2 + 
  ||g_v^2(h_v^2(\hat{i})) - p^{Contour}_v||_2^2 \\  
  &+||g_v^3(h_v^3(\hat{i})) - p^{RGB}_v||_2^2
  \end{split}
\label{loss_ventral}
\end{equation}
\begin{equation}
\begin{split}
  L_D &= 
  ||g_d^1(h_d^1(\hat{v})) - p^{Gabor}_d||_2^2 + 
  ||g_d^2(h_d^2(\hat{v})) - p^{Contour}_d||_2^2 \\
  &+||g_d^3(h_d^3(\hat{v})) - p^{Motion}_d||_2^2
  \end{split}
  \label{loss_dorsal}
\end{equation}
both $L_V$ and $L_D$ are weighted summation of the MSE losses at each intermediate blocks. $h_v$ and $h_d$ denote the feature extracting progress in two encoders (\eg, $h^2_v$ represents all the Transformer blocks up to the second intermediate block in ventral branch), $g_v$ and $g_d$ denote the decoders, $\hat{i}$ and $\hat{v}$ are masked images and masked videos, and $p$ denote the prediction targets of each branch. 

The objective of BIMM is a combination of two objectives:
\begin{equation}
  L =  L_V + \lambda L_D
\label{loss}
\end{equation}
$\lambda$ is a hyper-parameter that balances two branches.  In practice, we set $\lambda=1$.

\noindent\textbf{Pretraining.} 
Our pretraining progress can be divided into two stages.
First, the ventral branch is trained on ImageNet-1K \cite{russakovsky2015imagenet} by total loss function of Equation. \ref{loss_ventral}.
Then, we initialize the dorsal branch with the pretrained ventral model and train two branches jointly, utilizing the combined loss function as detailed in Equation. \ref{loss}. 
Detailed training progress is illustrated in Algorithm. \ref{pretrain_algorithmic}.


\noindent\textbf{Partial weight sharing strategy.} 
Inspired by the information sharing mechanism of V1 and V2 in human visual pathway, we design a partial weight sharing strategy between ventral and dorsal branch.
ViTs in both branches share the same network structure and multi-head attention mechanism \cite{arnab2021vivit, tong2022videomae}, thus simplifying the implementation of weight sharing. 
Specifically, we share the weights of first two intermediate blocks of each branch when pretrain BIMM jointly. 
Such a design not only maintains spatial knowledge learned from image datasets but also learns temporal information from video datasets.

\noindent\textbf{Finetuning and inference.} 
In the finetuning stage of BIMM, we utilize the pretrained position embedding layers along with the encoder from dorsal branch for video tasks and ventral branch for image tasks, respectively.
For adaptation to specific downstream tasks, we add task-specific layers to the trained model, \eg, adding an MLP for video action recognition tasks. 

\begin{algorithm}
\caption{The training strategy of BIMM.}
\begin{algorithmic}[1] 
    \Require Image dataset $I$, video dataset $V$ (\eg, K400), randomly initialized ViT models $f_{dorsal}$ and $f_{ventral}$. 
    \Ensure Pretrained ViT models from dorsal and ventral branches.
    \While{not end of ventral pretraining}
        \State \parbox[t]{0.9\linewidth}{Randomly select input images $i$ from $I$. Generate masked images $\hat{i}$ and progressive prediction targets $p_v'$.}
        \State \parbox[t]{0.9\linewidth}{Feed unmasked image patches to $f_{ventral}$.}
        \State \parbox[t]{0.9\linewidth}{Compute $L_V$ based on Equation. \ref{loss_ventral}.}
    \EndWhile
    \State Initialize the video encoder of dorsal branch $h_d$ with pretrained $h_v$.
    \While{not end of jointly pretraining}
        \State \parbox[t]{0.9\linewidth}{Randomly select input video $v$ from $V$. Generate masked videos $\hat{v}$ and progressive prediction targets $p_d$.}
        \State \parbox[t]{0.9\linewidth}{Randomly select masked video frames $\hat{f}$ from $\hat{v}$ and generate progressive prediction targets $p_v$.}
        \State \parbox[t]{0.9\linewidth}{Feed unmasked video patches to $f_{dorsal}$ and feed image patches to $f_{ventral}$.}
        \State \parbox[t]{0.9\linewidth}{Compute $L$ based on Equation. \ref{loss}.}
    \EndWhile
    \State Finetune $f_{ventral}$ on the test image dataset(\eg, ImageNet-1K or COCO or ADE20K).
    \State Finetune $f_{dorsal}$ on the test video datasets(\eg, UCF101 or SSv2 or K400).
\end{algorithmic}
\label{pretrain_algorithmic}
\end{algorithm}
\vspace{-2cm}

\section{Experiment}
\label{exp}
In this section, we conduct experiments on five popular video datasets (\ie, Kinetics-400 (K400) \cite{kay2017kinetics}, Something-Something-v2 (SSv2) \cite{goyal2017ssv2}, UCF101 \cite{ucf101}, HMDB51 \cite{hmdb51}, and AVA v2.2 \cite{gu2018ava}) and three image datasets (\ie, ImageNet-1K\cite{russakovsky2015imagenet}, COCO\cite{lin2014coco}, and ADE20K\cite{zhou2017ade20k}).
Diverse downstream tasks prove the effectiveness of our method, indicating both the dorsal and ventral branches learn comprehensive visual representations.
Besides, we carry out extensive ablation studies to investigate the significance of each components in our method.
The introduction of each dataset and implementation details are described in Appendix \textcolor{red}{A}.

\setlength{\tabcolsep}{8pt}
\begin{table*}[!t]
\centering
\caption{Comparison to the state-of-the-art on \textbf{K400}. \textcolor{gray}{Gray lines} denote supervised learning methods. \textbf{Bold numbers} indicate the best results under each test protocol. `$\star$' denotes the reproduced results. "N/A" indicates the numbers are not available. }
\vspace{-.3cm}
\resizebox{\linewidth}{!}{ 
\begin{tabu}{l|cccc|cc} 
\hline
Method & Backbone & Extra Data & Epoch & Frames & Top-1 & Top-5 \\ 
\hline
\textcolor{gray}{SlowFast\cite{feichtenhofer2019slowfast}} & \textcolor{gray}{ResNet101+NL} & \textcolor{gray}{no extra} & \textcolor{gray}{256} & \textcolor{gray}{16+64} & \textcolor{gray}{79.8} & \textcolor{gray}{93.9} \\
\textcolor{gray}{MViT-B\cite{fan2021mvit}} & \textcolor{gray}{MViT-B} & \textcolor{gray}{no extra}  & \textcolor{gray}{300} & \textcolor{gray}{32} & \textcolor{gray}{80.2} & \textcolor{gray}{94.4} \\
\textcolor{gray}{ViViT\cite{arnab2021vivit}} &\textcolor{gray}{ViViT-L} & \textcolor{gray}{JFT-300M}  & \textcolor{gray}{30} & \textcolor{gray}{128} & \textcolor{gray}{84.9} & \textcolor{gray}{95.8} \\
\textcolor{gray}{Video Swin\cite{liu2022videoswin}} &\textcolor{gray}{Swin-L} & \textcolor{gray}{IN-21K}  & \textcolor{gray}{30} & \textcolor{gray}{32} & \textcolor{gray}{83.1} & \textcolor{gray}{95.9} \\
\hline
BEVT\cite{wang2022bevt} & Swin-B & IN-21K+DALLE  & 1600+150 & 32 & 80.6 & N/A \\
OmniMAE\cite{girdhar2023omnimae} & ViT-B & IN-1K & 1600 & 16 &80.8 & N/A\\
\hline
\multirow{2}{*}{VideoMAE\cite{tong2022videomae}}& ViT-B & \multirow{2}{*}{no extra}  & \multirow{2}{*}{800} & \multirow{2}{*}{16} & 81.5 & 95.1 \\
& ViT-L &   &  &  & 85.2 & 96.8 \\
\hline
\multirow{2}{*}{MotionMAE\cite{yang2022motionmae}} & ViT-B & \multirow{2}{*}{no extra} & \multirow{2}{*}{1600} & \multirow{2}{*}{32} & 81.7 & N/A \\
& ViT-L &  &  &  & 85.3 & N/A \\
\hline
MAE-ST\cite{feichtenhofer2022maest}$^\star$ & ViT-B & \multirow{2}{*}{no extra} & \multirow{2}{*}{1600} & \multirow{2}{*}{16} & 80.6 & 94.7 \\
MAE-ST\cite{feichtenhofer2022maest} &ViT-L &  & &  & 84.8 & N/A \\
\hline
MaskFeat\cite{wei2022maskfeat}$^\star$ & MViT-B & \multirow{2}{*}{IN-1K} & \multirow{2}{*}{800} & \multirow{2}{*}{16} & 82.3 & 94.9 \\
MaskFeat\cite{wei2022maskfeat} & MViT-L & &  &  & 84.3 & 96.3 \\
\hline
\multirow{2}{*}{ \textbf{BIMM} }& ViT-B & \multirow{2}{*}{IN-1K}  &\multirow{2}{*}{800} & \multirow{2}{*}{16} & \textbf{85.0} & \textbf{95.8} \\
& ViT-L &  &  &  & \textbf{87.9} & \textbf{97.2} \\
\hline
\end{tabu}
}
\label{tab:sota_k400}
\vspace{-.5cm}
\end{table*}

\setlength{\tabcolsep}{8pt}
\begin{table*}[!thb]
\centering
\caption{Comparison to the state-of-the-art on \textbf{SSv2}.
The formatting in this table is consistent with the previous table.
}
\vspace{-.3cm}
\resizebox{\linewidth}{!}{ 
\begin{tabu}{l|cccc|cc}
\hline
Method & Backbone & Extra Data  & Epoch & Frames & Top-1 & Top-5 \\ 
\hline
\textcolor{gray}{ViViT\cite{arnab2021vivit}} &\textcolor{gray}{ViT-L} & \textcolor{gray}{no extra}  & \textcolor{gray}{35} & \textcolor{gray}{32} & \textcolor{gray}{65.9} & \textcolor{gray}{N/A} \\
\textcolor{gray}{SlowFast\cite{feichtenhofer2019slowfast}} & \textcolor{gray}{ResNet101} & \textcolor{gray}{K400} & \textcolor{gray}{256} & \textcolor{gray}{8+32} & \textcolor{gray}{63.1} & \textcolor{gray}{87.6} \\
\textcolor{gray}{MViT-B\cite{fan2021mvit}} & \textcolor{gray}{MViT-B} & \textcolor{gray}{K400}  & \textcolor{gray}{200} & \textcolor{gray}{64} & \textcolor{gray}{67.7} & \textcolor{gray}{90.9} \\
\textcolor{gray}{Video Swin\cite{liu2022videoswin}} & \textcolor{gray}{Swin-B} & \textcolor{gray}{IN-21K+K400}   & \textcolor{gray}{60} & \textcolor{gray}{32} & \textcolor{gray}{69.6} & \textcolor{gray}{92.7} \\
\hline
\multirow{4}{*}{VideoMAE\cite{tong2022videomae}}& ViT-B & no extra  & 2400 & 16 & 70.8 & 92.4 \\
 & ViT-L & no extra  & 2400 & 16 & 74.3 & 94.6\\
  & ViT-B & K400  & 800 & 16 & 69.7 & 92.3 \\
 & ViT-L & K400  & 800 & 16 & 74.0 & 94.6 \\
\hline
\multirow{3}{*}{MotionMAE\cite{yang2022motionmae}}& ViT-B & no extra  & 2400 & 16 & 71.8& N/A \\
 & ViT-L & no extra  & 2400 & 16 & 74.6 & N/A \\
 & ViT-L & K400  & 1600 & 16 & 74.3 & N/A \\
\hline
\multirow{3}{*}{OmniMAE\cite{girdhar2023omnimae}} & ViT-B & IN-1K  & 1600 & 16 & 69.5 & N/A\\
& ViT-L & IN-1K  & 1600 & 16 & 74.2 & N/A\\
& ViT-B & IN-1K+K400  & 1600 & 16 & 69.0 & N/A\\
\hline
MAE-ST\cite{feichtenhofer2022maest} &ViT-L & K400  & 1600 & 16 & 72.1 & N/A \\
MaskFeat\cite{wei2022maskfeat} & MViT-L & IN-1K+K400  & 800 & 16 & 73.3 & N/A \\
BEVT\cite{wang2022bevt} & Swin-B & IN-21K+K400+DALLE  & 1600+150 & 32 & 70.6 & N/A \\
\hline
\multirow{4}{*}{\textbf{BIMM}} & ViT-B & IN-1K  & 800 & 16 & \textbf{72.4} & \textbf{93.4} \\
 & ViT-L & IN-1K  & 800 & 16 & \textbf{75.1} & \textbf{96.0} \\
 & ViT-B & IN-1K+K400  & 800 & 16 & \textbf{72.7} & \textbf{93.8} \\
  & ViT-L & IN-1K+K400  & 800 & 16 & \textbf{74.6} & \textbf{96.0} \\
  \hline
\end{tabu}
}

\label{tab:sota_ssv2}
\vspace{-.8cm}
\end{table*}

\subsection{Main Results: Video Tasks}
We compare BIMM with the well-known Transformer-based methods in both supervised and self-supervised video learning domains. 
To ensure a comprehensive evaluation, we employ both domain-generic and domain-specific settings for SSv2, UCF101 and HMDB51. In the domain-generic setting, we pretrain and finetune BIMM on different datasets, whereas in the domain-specific setting, both pretraining and finetuning are conducted on the same dataset.

\noindent\textbf{K400.} K400 is a spatial heavily dataset. As shown in Table. \ref{tab:sota_k400}, BIMM achieves state-of-the-art performance among all self-supervised learning methods with ViT-B or ViT-L backbone, registering a Top-1 accuracy of 85.0\% and 87.9\%, respectively, which surpass the baseline \cite{tong2022videomae} by improvements up to 2.6\% and 2.7\%. 
Moreover, BIMM surpasses methods with stronger backbones, \eg, MViT-L in MaskFeat \cite{wei2022maskfeat}, or with extensive training data, \eg, BEVT \cite{wang2022bevt} pretrained with DALLE tokenizer \cite{ramesh2021dalle}. 
These results suggest that BIMM pretraining is more helpful for video action recognition.


\setlength{\tabcolsep}{8pt}
\begin{table}[!thb]
\caption{Comparison to the state-of-the-art on \textbf{AVA v2.2}. All models are pretrained in self-supervised manner on K400 at image size 224 $\times$ 224. 
The formatting of this table is consistent with the previous tables.
}
\vspace{-.2cm}
\centering
\begin{tabu}{l|cc|c} 
\hline
Method & Backbone &  Frames$\times$ sample rate  & mAP \\ 
\hline
$\rho$BYOL\cite{feichtenhofer2021rho} & SlowOnly-R50 &  $8\times 8$  & 23.4  \\
VideoMAE\cite{tong2022videomae} & ViT-B  & $16\times 4$   & 31.8  \\
VideoMAE\cite{tong2022videomae} & ViT-L  & $16\times 4$   & 34.3  \\
MAE-ST\cite{feichtenhofer2022maest} & ViT-L  & N/A  & 32.3  \\
MaskFeat\cite{wei2022maskfeat} & MViT-L & $40\times 3$  & 37.5 \\
\hline
\textbf{BIMM} & ViT-B & $16\times 4$  & \textbf{34.7}\\ 
\textbf{BIMM} & ViT-L & $16\times 4$  & \textbf{38.0}\\ 
\hline
\end{tabu}
\label{tab:sota_ava}
\vspace{-.7cm}
\end{table}

\setlength{\tabcolsep}{8pt}
\begin{table*}[!thb]
\centering
\caption{Computing cost comparing with popular masked video modeling methods. All methods are pretrained and finetuned on K400.}
\vspace{-.2cm}
\begin{tabu}{l|ccc|c} 
\hline
Method & Backbone &  GFLOPs & Parameters & Top-1 \\ 
\hline
BEVT\cite{wang2022bevt} & Swin-B & $282\times4\times3$ & 88M & 80.6 \\
VideoMAE\cite{tong2022videomae} & ViT-B & $180\times5\times3$ & 87M & 81.5 \\
VideoMAE\cite{tong2022videomae} & ViT-L & $597\times5\times3$ & 305M & 85.2 \\
MotionMAE\cite{yang2022motionmae} & ViT-B & $180\times5\times3$ & 87M & 81.7 \\
MotionMAE\cite{yang2022motionmae} & ViT-L  & $598\times5\times3$ & 305M & 85.3 \\
MAE-ST\cite{feichtenhofer2022maest}$^\star$& ViT-B  & $180\times5\times3$ & 87M & 80.6  \\
MaskFeat\cite{wei2022maskfeat} & MViT-L  &$377\times10\times1$ & 218M & 84.3  \\
\hline
\textbf{BIMM} & ViT-B  &$180\times5\times3$ & 87M & \textbf{85.0}  \\
\textbf{BIMM} & ViT-L  &$597\times5\times3$ & 305M & \textbf{87.9}  \\
\hline
\end{tabu}
\label{tab:GFLOPs_sota}
\vspace{-1cm}
\end{table*}

\noindent\textbf{SSv2.} Contrary to the spatially-heavy K400, SSv2 is more sensitive to temporal information. According to Table. \ref{tab:sota_ssv2}, BIMM using ViT-B backbone achieves Top-1 accuracies of 72.7\% and 72.4\% under domain-generic and domain-specific settings, which surpasses all other supervised and self-supervised learning methods.
In addition, when using ViT-L, BIMM achieves the state-of-the-art performance in both settings with 74.6\% and 75.1\% in terms of Top-1 accuracy.
Such results indicate that BIMM is also good at capturing temporal information.
It is noteworthy that, compared to other methods that use the ViT-B backbone for pretraining on the SSv2, our method attains superior learning outcomes with a notable reduction in the number of required training epochs

\noindent\textbf{UCF101 and HMDB51.} 
Our detailed comparison with the state-of-the-art methods on UCF101 and HMDB51 is showcased in Appendix \textcolor{red}{B}.
When pretrained on K400, BIMM achieves 97.2\% on UCF101 and 76.3\% on HMDB51 in terms of Top-1 accuracy.
When pretrained on UCF101, our method also showcases the superiority with 94.8\% on UCF101 and 65.2\% on HMDB51 in terms of Top-1 accuracy.
Furthermore, utilizing a more powerful backbone, ViT-L, enhances the performance of BIMM on both UCF101 and HMDB51 datasets during the inference phase.

\noindent\textbf{AVA.} 
Table. \ref{tab:sota_ava} reports mean Average Precision (mAP) of action detection on AVA dataset. All models are pretrained on K400 with the input size of $224\times224$. 
BIMM, when using ViT-B as the backbone, achieves a significant improvement of 2.9 in terms of mAP over the baseline VideoMAE, demonstrating its capability for more comprehensive video representation learning. 
Similarly, when equipped with ViT-L, BIMM surpasses all existing masked modeling methods, including MaskFeat \cite{wei2022maskfeat}, despite the latter sampling more video frames for evaluation.

\noindent\textbf{Computational cost.}
To compare computational costs, we enumerate the GFLOPs and parameters of popular masked video modeling methods in Table. \ref{tab:GFLOPs_sota}. 
Following the common practices in \cite{tong2022videomae,yang2022motionmae,wei2022maskfeat}, 
Table. \ref{tab:GFLOPs_sota} reports the GFLOPs and parameters during the inference phase of each model. Methods using the same backbone obtain similar computing costs.
As can be seen, BIMM achieves a significant improvement of Top-1 accuracy with similar computational cost to other methods on K400. 

\begin{table}[!thb]
\centering
\begin{minipage}[t]{.49\linewidth}
\centering
\caption{\textbf{ImageNet-1K image classification}. 
All models are pretrained and finetuned on ImageNet-1K under $224 \times 224$ resolution and using ViT-B backbone.}
\vspace{-.2cm}
\setlength{\tabcolsep}{8pt}
\resizebox{\linewidth}{!}{
\begin{tabu}{l|ccc} 
\hline
Method & Epochs &  Target & Top-1 Acc (\%) \\ 
\hline
BeiT\cite{bao2021beit}& 800 & DALL-E &  83.2 \\ 
SimMIM\cite{xie2022simmim}& 800 & Pixel & 83.8 \\
MaskFeat\cite{wei2022maskfeat}& 1600& HOG&84.0 \\
PeCo\cite{dong2023peco}&800 & CodeBook&84.5 \\
MAE\cite{he2022mae}&1600 & Pixel& 83.6\\
DeepMIM\cite{ren2023deepmim}& 1600&Pixel & 84.0\\
LocalMIM\cite{wang2023localmim}& 1600& HOG&84.0 \\
\hline
BIMM-V & 800 & Progressive & 84.8\\
BIMM & 800+1600 & Progressive & \textbf{85.2}\\
\hline
\end{tabu}
}
\label{tab:sota-IN1K}
\end{minipage}%
\hspace{.3cm}
\begin{minipage}[t]{.45\linewidth}
\centering
\caption{\textbf{COCO object detection and ADE20K semantic segmentation}. 
For COCO, we report 
box AP (AP$^{box}$) for object detection and mask AP (AP$^{mask}$) for instance segmentation. 
For ADE20K, we report the mIoU of each method.}
\vspace{-.2cm}
\setlength{\tabcolsep}{8pt}
\resizebox{\linewidth}{!}{
\begin{tabu}{l|cc|c} 
\hline
\multirow{2}{*}{Method} & \multicolumn{2}{c|}{COCO} & ADE20K\\
\cline{2-4}
 & AP$^{box}$ &  AP$^{mask}$ & mIoU \\ 
\hline
BeiT\cite{bao2021beit}& 49.8 & 44.4 & 47.1\\ 
SimMIM\cite{xie2022simmim}& 49.1 & 43.8 & N/A \\
PeCo\cite{dong2023peco}& 44.9 & 40.4 & 48.5 \\
MAE\cite{he2022mae}& 50.3 & 44.9 & 48.1\\
DeepMIM\cite{ren2023deepmim}& 51.6 & 45.2 & 49.5\\
LocalMIM\cite{wang2023localmim}& 50.7 & 44.9 & 49.5 \\
\hline
BIMM & \textbf{51.8} & \textbf{45.5} & \textbf{50.3}\\
\hline
\end{tabu}
}

\label{tab:sota-COCO&ADE20K}
\end{minipage}
\vspace{-1.0cm}
\end{table}

\subsection{Main Results: Image Tasks}
In this section, we compare the performance of BIMM with other popular masked image modeling methods on three downstream image tasks.

\noindent\textbf{Image classification.} 
In Table. \ref{tab:sota-IN1K}, we compare the finetuning results of various self-supervised pretraining methods on ImageNet-1K. For this experiment, we pretrain the ViT-B in ventral branch (denoted as BIMM-V in Table.\ref{tab:sota-IN1K}) for 800 epochs on ImageNet-1K in an unsupervised manner and then finetune it for 100 epochs. Finally, we report the accuracy on the validation set.
BIMM-V achieves the top-1 accuracy of 84.8\%, surpassing the baseline MAE \cite{he2022mae}, as well as the more advanced counterparts DeepMIM \cite{ren2023deepmim} and LocalMIM \cite{wang2023localmim}.
Furthermore, we also test the ViT-B from dorsal branch after jointly training two branches of BIMM on K400 (denoted as BIMM in Table. \ref{tab:sota-IN1K}). We are pleased to observe that training on video datasets can lead to further performance improvement.

\noindent\textbf{Object detection.} 
Following common practice, we use the COCO benchmark \cite{lin2014coco} to evaluate the transferability of BIMM to the object detection task. We employ pretrained ViT-B as the backbone and Mask R-CNN\cite{he2017mask} as the detector. We report box Average Precision (AP$^{box}$) for object detection and mask Average Precision (AP$^{mask}$) for instance segmentation in Table.\ref{tab:sota-COCO&ADE20K}. BIMM outperforms the MAE baseline by 1.5 in terms of AP$^{box}$ and 0.6 in terms of AP$^{mask}$, respectively. 

\begin{table*}[!ht]
\makebox[\textwidth][c]{\begin{minipage}{\linewidth}
\caption{\textbf{Ablation studies}. "Acc" denotes the Top-1 accuracy of action recognition task on UCF101. 
The entries marked in \colorbox{gray!30}{gray} are the same, which specify the default settings.
\label{tab:ablations}
}
\vspace{-.2cm}
\centering
\subfloat[
    \textbf{Initialization of dorsal branch}. Initialized by ventral branch pretrained on IN-1K works the best.
    \label{tab:4_table}
]{
\begin{minipage}{0.26\linewidth}{\begin{center}
\fontsize{8pt}{10pt}\selectfont
    \begin{tabular}{lc}
        \hline
        Initialization & Acc(\%) \\
        \hline
        No Init & 90.9 \\
       UCF101 Init& 91.2 \\
        IN-1K Init& \colorbox{gray!30}{\textbf{92.5}}\\
        \end{tabular}
\end{center}}\end{minipage}
}
\hspace{1em}
\subfloat[
    \textbf{Pretrain without ventral branch}. Without the ventral branch, performance of BIMM decreases, but still higher than the baseline.
    \label{tab:_table}
]{
\begin{minipage}{0.26\linewidth}{\begin{center}
\fontsize{8pt}{10pt}\selectfont
    \begin{tabular}{lc}
        \hline
        Method & Acc(\%) \\
        \hline
        Baseline & 90.7 \\
        BIMM-Dorsal & 91.0 \\
        BIMM & \colorbox{gray!30}{\textbf{92.5}}\\
        \end{tabular}
\end{center}}\end{minipage}
}
\hspace{1em}
\subfloat[
    \textbf{Ablation of prediction target}. Each prediction target contributes to the representation learning.
    \label{tab:six_table}
]{
\centering
\begin{minipage}{0.3\linewidth}{\begin{center}
\fontsize{8pt}{10pt}\selectfont
    \begin{tabular}{lc}
        \hline
        Prediction target  & Acc(\%) \\
        \hline
        V1 & 89.3 \\
        V1-V2 & 90.2\\
        V1-V2-V4\&MT & \colorbox{gray!30}{\textbf{92.5}}\\ 
        \end{tabular}
\end{center}}\end{minipage}
}
\\
\subfloat[
    \textbf{Separation of intermediate blocks}. The separation of 2nd, 4th, and 12th yields the best result.
    \label{tab:first_table}
]{
\begin{minipage}{0.26\linewidth}{\begin{center}
\fontsize{8pt}{10pt}\selectfont
    \begin{tabular}{lc}
        \hline
        Separation &  Acc(\%) \\
        \hline
        12 &  91.0\\
        4-8-12 &  91.7\\
        6-9-12 &  91.3\\
        2-4-12 &  \colorbox{gray!30}{\textbf{92.5}}\\
        \end{tabular}
\end{center}}\end{minipage}
}
\hspace{1em}
\subfloat[
    \textbf{Partial weight sharing}. Sharing the entire Transformer network or not share any weights leads to declines in performance.
    \label{tab:second_table}
]{
\centering
\begin{minipage}{0.26\linewidth}{\begin{center}
\fontsize{8pt}{10pt}\selectfont
    \begin{tabular}{lc}
        \hline
        Sharing strategy &  Acc(\%) \\
        \hline
        No sharing & 91.0 \\
        Partial Sharing  & \colorbox{gray!30}{\textbf{92.5}}\\
        All Sharing & 91.3\\
        ~\\
        \end{tabular}
\end{center}}\end{minipage}
}
\hspace{1em}
\subfloat[
    \textbf{Mask ratio}. Masking 90\% video patches works the best. The mask ratio of ventral branch keeps 0.75.
    \label{tab:third_table}
]{
\centering
\begin{minipage}{0.30\linewidth}{\begin{center}
\fontsize{8pt}{10pt}\selectfont
    \begin{tabular}{ccc}
        \hline
         Mask ratio  & Acc(\%) \\
        \hline
        0.5  &91.7\\
        0.75  &91.9\\
        0.9 & \colorbox{gray!30}{\textbf{92.5}}\\
        0.95  &91.5\\
        \end{tabular}
\end{center}}\end{minipage}
}
\\
\vspace{-.4cm}
\end{minipage}}
\vspace{-1.2cm}
\end{table*}

\noindent\textbf{Semantic segmentation.} 
We also transfer the pretrained ViT-B to semantic segmentation on the ADE20K benchmark \cite{zhou2017ade20k}. We use UperNet\cite{xiao2018upernet} for a fair comparison with previous methods. Mean intersection over union (mIoU) for each model is reported in Table.\ref{tab:sota-COCO&ADE20K}. BIMM surpasses its most advanced counterpart by 0.8 mIoU.

These experiments indicate that the ventral branch of BIMM is capable of learning good spatial representations, contributing to the performance of downstream image tasks and the optimization of the dorsal branch.

\subsection{Ablation Studies}
We provide a set of ablation studies to justify the contribution of different components in BIMM. If not specially mentioned, all experiments in this section are based on pretraining our framework using ViT-B backbone on UCF101 for 800 epochs, followed by a finetuning phase of 100 epochs on the same dataset. Studies about training schedule and more visualization results are provided in Appendix \textcolor{red}{B}.

\noindent\textbf{Initialization of dorsal branch.}
In the pretraining of BIMM, we first pretrain ventral branch individually on the ImageNet-1K, which efficiently acquires spatial representations. 
This pretrained model then serves as the initialization of the ViT in the dorsal branch. 
Therefore, we conducted experiments under other two settings: one trains the dorsal branch from scratch without initialization, another initializes the dorsal branch with the ventral branch pretrained on the images from UCF101.
The results are reported in Table. \ref{tab:4_table}.
Comparing to the default setting, the performance falls without the initialization from ventral branch.
Additionally, when the ventral branch is pretrained on the images from UCF101, there is still a slight decrease of 1.3\% in Top-1 accuracy on action recognition.
This study demonstrates that the static image information learned by the ventral branch can be beneficial to the overall video representation learning.


\noindent\textbf{Pretrain without ventral branch.}
We pretrain the dorsal branch (BIMM-Dorsal) independently on a video dataset, without any initialization or jointly training with the ventral branch.
The results of this experiment, as depicted in Table. \ref{tab:_table}, indicate that even in the absence of the ventral branch and the partial weight sharing strategy, BIMM-Dorsal manages to surpass the VideoMAE baseline. 
It is worth noting that the performance reported for VideoMAE are based on its pretraining for 1600 epochs on UCF101.
This finding highlights the effectiveness of the intermediate blocks and progressive prediction targets.

\noindent\textbf{Progressive prediction targets.}
In light of the visual pathway in human brain, the core design of BIMM is the progressive prediction targets. 
We hope to further explore whether the combination of prediction targets is optimal. 
Therefore, we perform ablation on each prediction target, gradually increasing the prediction targets over three experiments.
For example, "V1" indicates during the joint training progress, we only calculate the loss from the V1 blocks, which predict the Gabor feature.
As reported in Table. \ref{tab:six_table}, BIMM using complete progressive prediction targets performs the best. Reducing any of the prediction targets results in a decline in downstream task accuracy.

\noindent\textbf{Separation of intermediate blocks.}
Each branch of BIMM are divided into multiple intermediate blocks to learn from the prediction targets. Therefore, the separation of blocks is worth exploring.
In this study, we simultaneously adjust the distribution of intermediate blocks in both the ventral and dorsal branches. As indicated in Table. \ref{tab:first_table}, we found that a configuration with intermediate blocks separated at the 2nd, 4th, and 12th Transformer blocks delivers the best result, surpassing the VideoMAE baseline(90.7\%) by a margin of 1.8\%.

\noindent\textbf{Partial weight sharing.}
By default, we share the parameters of the first two intermediate blocks between two branches during joint training.
However, we wonder how the partial weight sharing strategy contributes to the training.
Therefore, we conduct ablation experiments under two settings: firstly, no parameters are shared between the two branches (denoted as "No sharing"); secondly, all parameters of the Transformer blocks within the two branches are shared  (denoted as "All Sharing").
As shown in Table. \ref{tab:second_table}, 
partial weight sharing strategy (denoted as "Partial Sharing") performs the best, indicating the difference between intermediate blocks and the effectiveness of the strategy.

\noindent\textbf{Mask ratio.}
In this study, we investigate the optimal mask ratio in the dorsal branch.
As seen in Table. \ref{tab:third_table}, we conduct experiments with mask ratios ranging from 0.5 to 0.95 and observe the optimal performance is obtained at a mask ratio of 0.9.
It is worth noting that mask ratio of ventral branch is maintained at 0.75 for all experiments, consistent with MAE.

\section{Limitation} 

BIMM draws inspiration from the visual information processing progress in the visual pathway in human brain, and aims to incorporate such neuroscience knowledge into masked modeling methods, enhancing the ability to learn visual representations. 
However, many operational mechanisms of the human brain are still mysterious.
Applying other neuroscience knowledge, involving the function of V3 area, bidirectional links and cross-layer connections in visual pathway, to self-supervised learning methods still deserves further exploration.

\section{Conclusion}

In this study, we propose the Brain Inspired Masked Modeling (BIMM) framework, which conducts self-supervised video representation learning inspired by the process of visual information processing in the human brain. 
Drawing inspiration from the ventral and dorsal pathways of the visual cortex, BIMM employs a dual-branch structure, which incorporates progressive prediction targets and a partial weight sharing strategy.
This design enables the framework to simultaneously process static and dynamic visual information and efficiently capture a wide array of visual features, ranging from basic textures and contours to complex motion patterns and high-level semantic content.
Each branch of BIMM has a ViT encoder divided into three intermediate blocks attached with lightweight decoders, allowing the model to learn richer and more comprehensive representations.
Our experiments on various datasets demonstrated that BIMM outperforms current state-of-the-art methods in both video and image tasks.

%
%
\bibliographystyle{splncs04}
\bibliography{main}

\clearpage
\setcounter{page}{1}
\appendix

\section*{Appendix}

\section{Implementation Details}
\label{sec:appendix_imple}
All experiments are conducted on NVIDIA GeForce RTX 4090 GPUs, using PyTorch \cite{paszke2019pytorch} 1.10.0 and CUDA 11.3. 
As for training strategy, we first train the ventral branch on ImageNet-1K for 800 epochs and then jointly train ventral and dorsal branches on each video dataset.
In this section, we briefly introduce the datasets and report the implementation details for pretraining and finetuning on each of them.
For simplicity, we list the learning rate, batch size and other necessary hyperparameters in Table. \ref{tab:pretrain_config} and Table. \ref{tab:finetune_config}.

\subsection{Image Datasets}

\noindent\textbf{ImageNet-1K(IN-1K)\cite{russakovsky2015imagenet}.}
IN-1K is a popular image dataset containing approximately 1.2 million images categorized into 1,000 different classes.
The dataset is released under a non-commercial license and this subset of ImageNet is widely used for benchmarking image recognition models.
The pretraining and finetuning settings of IN-1K mostly follow \cite{he2022mae, wang2023localmim}. 
The input resolution for both pretraining and finetuning are $224 \times 224$.
We pretrain the ventral branch on IN-1K for 800 epochs and jointly train it with dorsal branch. During single pretraining progress, we use simple data augmentation and apply linear learning rate scaling rule: $lr = baselr \times batchsize/16$. 
During joint pretraining of both branches, the batch size for ventral branch is kept the same as for the dorsal branch. 
For inference, we report the Top-1 classification accuracy on the validation set. 

\noindent\textbf{COCO\cite{lin2014coco}.}
COCO is a dataset widely-used in object detection task, which contains photos of 91 objects types with a total of 2.5 million labeled instances in 328k images. 
After joint pretraining of two branches in BIMM, we use the ViT-B from ventral branch as the backbone. Following \cite{he2022mae,ren2023deepmim}, the model is finetuned on COCO train split and evaluated on validation split. We train the model for a total of 36 epochs and decays the learning rate by a factor of 10 at the 27$th$ and 33$rd$ epochs. We use AdamW optimizer with the learning rate of $1e^{-4}$ and weight decay of 0.05.

\noindent\textbf{ADE20K\cite{zhou2017ade20k}.}
ADE20K contains over 20K images with dense annotations, comprising 430K objects from more than 2,600 categories, making it a popular dataset for semantic segmentation.
After joint pretraining of two branches in BIMM, we use the ViT-B from ventral branch as the backbone. Following \cite{he2022mae, wang2023localmim}, we finetune end-to-end for 160K iterations using AdamW optimizer with the peak learning rate of $4e^{-4}$, weight decay of 0.05 and batch size of 16. The learning rate warms up over 1500 iterations and then decays with linear strategy. The model is trained with the input resolution of $512\times512$ and uses bilinear positional embedding to interpolate.

\subsection{Video Datasets}

\noindent\textbf{Kinetics-400(K400)\cite{kay2017kinetics}.}
K400 is a large-scale dataset containing 246K training videos and 20K validation videos spanning 400 categories. 
The dataset, based on publicly available web videos from YouTube. Due to the videos being taken down over time, the dataset changes over time making apples-to-apples comparison with prior work difficult.
Hence, we use the static dataset provided by Opendatalab \footnote{https://opendatalab.com/Kinetics-400} following VideoMAE.
To the best of our knowledge, no PII or harmful content has been reported on this dataset.
We pretrain BIMM on K400 for 800 epochs with a cosine decay schedule, and then use a linear warmup strategy for the first 40 epochs. We extract the class token after the last stage and use it as the input to the final linear layer to predict the output classes.  
During finetuning, we adopt dense sampling following Slowfast \cite{feichtenhofer2019slowfast}. 
For detailed settings, please refer to Table. \ref{tab:pretrain_config} and Table. \ref{tab:finetune_config}.

\noindent\textbf{SSv2\cite{goyal2017ssv2}.}
SSv2, another extensive video dataset, emphasizes temporal information with its 169K training and 20K validation videos across 174 action classes. 
The dataset has been collected by consenting participants who recorded the videos given the action label, and released under a non-commercial license. To the best of our knowledge, no PII or harmful content has been reported in the dataset.
BIMM is pretrained for 800 epochs on SSv2 by default. During the finetuning stage, we perform the uniform sampling following \cite{wang2018temporal}. 
For detailed settings, please refer to Table. \ref{tab:pretrain_config} and Table. \ref{tab:finetune_config}.


\setlength{\tabcolsep}{8pt}
\begin{table*}[t]
\centering
\resizebox{0.6\linewidth}{!}{
\begin{tabu}{l|cccc} \hline
Config & K400 &  SSv2 & UCF &IN-1K \\ 
\hline
optimizer & AdamW & AdamW& AdamW& AdamW\\
base lr& 1.5e-4 & 1.5e-4  & 3e-4 &2e-4 \\ 
weight decay& 0.05& 0.05& 0.05& 0.05\\ 
\multirow{2}{*}{optimizer momentum}&$\beta_1 = 0.9$ &$\beta_1 = 0.9$&$\beta_1 = 0.9$&$\beta_1 = 0.9$\\ 
&$\beta_2 = 0.95$&$\beta_2 = 0.95$&$\beta_2 = 0.95$&$\beta_2 = 0.95$\\
batch size& 16 & 16  & 64 & 128 \\ 
learning rate schedule& cos decay& cos decay& cos decay& cos decay \\ 
warmup epochs& 40& 40& 40& 40 \\ 
random flip& no & yes &  yes &no\\ 
random crop& yes& yes& yes& yes  \\ 
\hline
\end{tabu}
} 
\caption{\textbf{Pretraining settings}. ``IN-1K'' denotes the pretrain setting of ventral branch. When singly training ventral branch, the batch size is 128. When training branches jointly, the batch size is same as the corresponding video dataset.}
\vspace{-0.8cm}
\label{tab:pretrain_config}
\end{table*}

\noindent\textbf{UCF101\cite{ucf101}.}
UCF101 comprises 13,320 videos covering 101 action classes.
We pretrain BIMM for 1600 epochs on UCF101. Here, 16 frames with a temporal stride of 4 are sampled. For finetuning, the model is trained with repeated augmentation \cite{hoffer2020augment}. 
For detailed settings, please refer to Table. \ref{tab:pretrain_config} and Table. \ref{tab:finetune_config}.

\noindent\textbf{HMDB51\cite{hmdb51}.}
HMDB51 includes 6,766 videos in 51 classes. 
We do not pretrain BIMM on HMDB51, only finetune other pretrained models on it. Here, 16 frames with a temporal stride of 2 are sampled. The augmentation during finetuning is consistent with UCF101. 
For detailed settings, please refer to Table. \ref{tab:finetune_config}.

\noindent\textbf{AVA v2.2\cite{gu2018ava}.}
AVA v2.2 is a dataset for spatiotemporal action localization, which contains the bounding box annotations and the corresponding action labels on keyframes. It has 211k training and 57k validation video segments.
We follow the standard protocol reporting mean Average Precision (mAP) on 60 classes on AVA v2.2.
Following the action detection manner in Slowfast \cite{feichtenhofer2019slowfast}, we resize original videos from the resolution of $450\times 360$ to $320\times 256$. During training, we apply random crop to $224\times 224$ and random flip as augmentation.
For training, we initialize the network weights from the model pretrained on K400 and use the ground-truth human detection boxes as training samples. For detailed settings, please refer to Table. \ref{tab:finetune_config}.
For inference, we perform inference on a single clip with 16 frames sampled with stride 4 centered at the frame, and use the detected person boxes with confidence more than 0.8 from AIA \cite{tang2020asynchronous}.

\noindent\textbf{License of Video Data}
All the datasets we used are commonly used datasets for academic purpose. The license of the Something-Something v2\footnote{https://developer.qualcomm.com/software/ai-datasets/something-something} and UCF101\footnote{https://www.crcv.ucf.edu/data/UCF101.php} datasets is custom. The license of the Kinetics-400\footnote{https://www.deepmind.com/open-source/kinetics}, HMDB51\footnote{https://serre-lab.clps.brown.edu/resource/hmdb-a-large-human-motion-database} and AVA\footnote{https://research.google.com/ava/index.html} datasets is CC BY-NC 4.0\footnote{https://creativecommons.org/licenses/by/4.0}.


\setlength{\tabcolsep}{8pt}
\begin{table*}[t]
\centering
\resizebox{\linewidth}{!}{
\begin{tabu}{l|cccccc} \hline
Config & K400 &  SSv2 & UCF & HMDB51 & AVA &IN-1K\\ 
\hline
optimizer &AdamW&AdamW&AdamW&AdamW&AdamW&AdamW \\ 
base lr& 1e-3 &  5e-4 & 5e-4 &  1e-3 & 2.5e-4 & 2e-3\\ 
weight decay& 0.05& 0.05& 0.05& 0.05& 0.05& 0.05 \\ 
\multirow{2}{*}{optimizer momentum}&  $\beta_1 = 0.9$&  $\beta_1 = 0.9$&  $\beta_1 = 0.9$&  $\beta_1 = 0.9$&  $\beta_1 = 0.9$&  $\beta_1 = 0.9$ \\ 
&$\beta_2 = 0.999$&$\beta_2 = 0.999$&$\beta_2 = 0.999$&$\beta_2 = 0.999$&$\beta_2 = 0.999$&$\beta_2 = 0.999$ \\
batch size& 16 & 16  & 32 & 32  & 16 & 128\\ 
learning rate schedule& cos decay & cos decay & cos decay & cos decay & cos decay & cos decay  \\ 
warmup epochs& 5&5&5&5&5 & 20 \\ 
training epochs& 75(B) 50(L) &  40(B), 30(L) & 100 &  50 & 30 & 100 \\ 
evaluation protocol (clips $\times$ crops) &$5\times 3$&$2\times 3$&$5\times 3$&$10\times 3$&-&-\\
repeated augmentation& 2 & 2  & 2 & 2  & no & no \\ 
random flip& yes &  no & yes & yes  &  yes & no\\ 
label smoothing&0.1 &0.1 &0.1 &0.1 &0.1 &0.1 \\ 
mixup& 0.8& 0.8& 0.8& 0.8& 0.8& 0.8\\ 
cutmix&  1.0&  1.0&  1.0&  1.0&  1.0&  1.0\\ 
drop path& 0.1(B), 0.2(L) & 0.1(B), 0.2(L)  & 0.2 & 0.2  & 0.2 & 0.1 \\ 
layer-wise lr decay& 0.75 & 0.75  & 0.7 & 0.7  & 0.75 & 0.65\\ 
\hline
\end{tabu}
}
\caption{\textbf{Finetuning settings}. For the video recognition and action detection downstream tasks, we finetune the pretrained ViT in the dorsal branch. For image classification downstream task, we use the pretrained network in the ventral branch. (B) denotes the parameter for ViT-B, (L) denotes the parameter for ViT-L.}
\vspace{-0.8cm}
\label{tab:finetune_config}
\end{table*}

\section{Additional Results}
\label{sec:appendix_result}

\subsection{Comparison with the State-of-the-art Methods}
\noindent\textbf{Action recognition on UCF101 and HMDB51.} 
Our detailed comparison with current state-of-the-art methods on UCF101 and HMDB51 is showcased in Table. \ref{tab:sota_ucf}. 
To further emphasize the effectiveness of BIMM, we compare with some popular contrastive learning methods. 
When pretrained on K400 with ViT-B, BIMM achieves Top-1 accuracy of 97.2\% on UCF101 and 76.3\% on HMDB51, surpasses the previously best contrastive learning method, $\rho$BYOL \cite{feichtenhofer2021rho}, by 3.0\% and 4.2\%, respectively. 
When pretrained on UCF101, our method also exhibits the superiority and achieves the best performance comparing with previous methods.
Last but not least, using more powerful ViT-L as the backbone can further enhance the performance of downstream tasks.


\subsection{Ablation study} 

\begin{figure*}[htbp]
    \centering
    \begin{subfigure}[b]{\textwidth}
        \includegraphics[width=\textwidth]{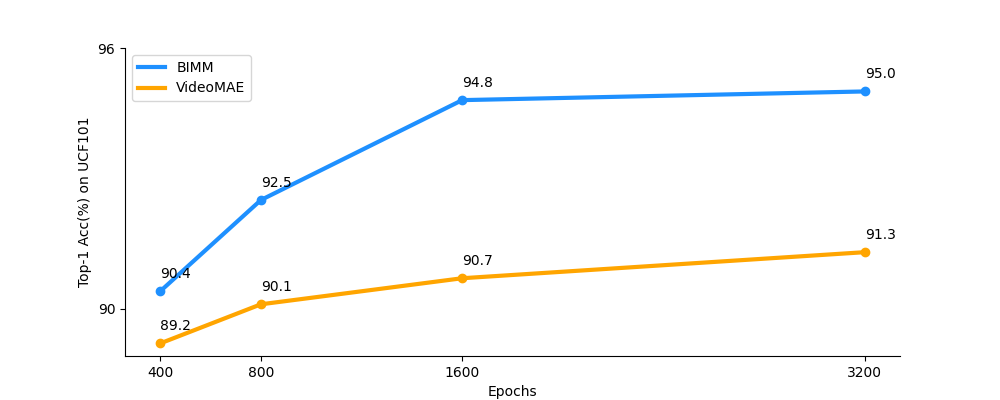}
        \caption{We pretrain BIMM on UCF101 for different epochs and report the Top-1 finetuning accuracy on UCF101 action recognition task. The results of VideoMAE are reproduced through its released code.}
        \label{fig:ucf101}
    \end{subfigure}
    \hfill
    \begin{subfigure}[b]{\textwidth}
        \includegraphics[width=\textwidth]{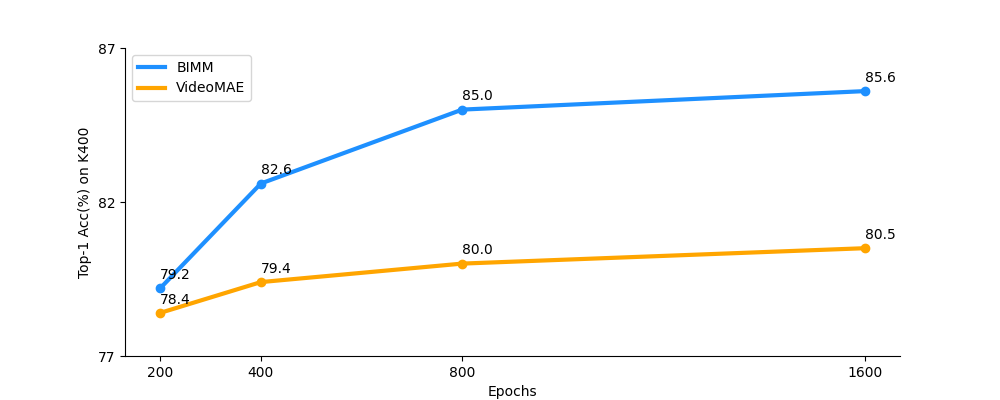}
        \caption{We pretrain BIMM on K400 for different epochs and report the Top-1 finetuning accuracy on K400 action recognition task. The results of VideoMAE are obtained from its paper.}
        \label{fig:k400}
    \end{subfigure}
    \caption{Ablation on \textbf{training schedule}. After training for 800 epochs on K400 and 1600 epochs on UCF101, longer pretraining epochs do not lead to significant improvement. Other settings keep the same as the default.}
    \label{fig:training_schedule}
\end{figure*}

\noindent\textbf{Training schedule.}
Figure. \ref{fig:training_schedule} shows the influence of the longer pretraining schedule on the K400 and UCF101 datasets. We find that a longer pretraining schedule brings slight gains to both datasets. Therefore in the main paper, BIMM is pretrained for 800 epochs on K400 and 1600 epochs on UCF101, respectively. 
However, the ablation studies are conducted with 800 epochs of pretraining on UCF101, for saving time and reducing computational cost.

\begin{table*}[t]
\centering
\resizebox{\linewidth}{!}{
\begin{tabu}{l|ccccc|c|c} \hline
Method & Backbone & Pretrain Data  & Epoch & Frames & Parameters & UCF101 & HMDB51 \\ 
\hline
OPN\cite{lee2017unsupervised}& VGG & UCF101 & N/A & 16  & 5.8M & 59.6 & 23.8 \\
VCOP\cite{xu2019self}& R(2+1)D & UCF101 & 300 & 16 & N/A & 72.4 & 30.9 \\
CoCLR\cite{han2020self}& S3D-G & UCF101  & 700 & 32  & 9M & 81.4 & 52.1 \\
Vi$^2$CLR \cite{diba2021vi2clr}& S3D & UCF101  & 300 & 32  & 9M & 82.8 & 52.9 \\
VideoMAE\cite{tong2022videomae} & ViT-B & UCF101  & 3200 & 16  & 87M & 91.3 & 62.6 \\
MotionMAE\cite{yang2022motionmae} & ViT-B & UCF101 & 2400 & 16  & 87M & 94.0 & N/A \\
\hline
BIMM & ViT-B & IN-1K+UCF101 & 1600 & 16 & 87M & \textbf{94.8} & \textbf{65.2} \\
BIMM & ViT-L & IN-1K+UCF101 & 1600 & 16 & 305M & \textbf{97.6} & \textbf{67.1} \\
\hline
VideoMoCo\cite{pan2021videomoco}&  R(2+1)D  & K400  & 200 & 16  & 15M & 78.7 & 49.2 \\
CoCLR\cite{han2020self}& S3D-G & K400  & 500 & 32  & 9M & 87.9 & 54.6 \\
Vi$^2$CLR\cite{diba2021vi2clr}& S3D & K400  & 300 & 32  & 9M & 89.1 & 55.7 \\
$\rho$BYOL\cite{feichtenhofer2021rho}& SlowOnly-R50 & K400  & 200 & 8 & 32M & 94.2 & 72.1 \\
VideoMAE\cite{tong2022videomae}& ViT-B & K400  & 800 & 16 & 87M & 96.1 & 73.3 \\
MotionMAE\cite{yang2022motionmae} & ViT-B & K400  & 1600 & 16  & 87M & 96.3 & N/A \\
\hline
BIMM & ViT-B & IN-1K+K400  & 800 & 16 & 87M & \textbf{97.2} & \textbf{76.3} \\
BIMM & ViT-L & IN-1K+K400  & 800 & 16 & 305M & \textbf{98.5} & \textbf{78.0} \\
\hline 
\end{tabu}
}
\caption{Top-1 finetuning accuracy on \textbf{UCF101 \& HMDB51} action recognition tasks. ``N/A" indicates the numbers are not available.}
\vspace{-0.8cm}
\label{tab:sota_ucf}
\end{table*}

\subsection{Qualitative Results}
\label{appendix:qualitative}
\begin{figure}[ht]
    \centering
    \begin{subfigure}{\textwidth}
        \centering
        \includegraphics[width=0.8\textwidth]{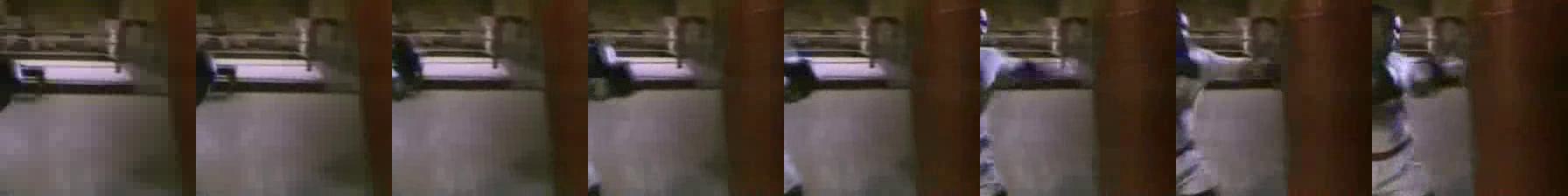}
        \caption*{GT: Punch\\ \textcolor{green}{BIMM: Punch}, \textcolor{red}{VideoMAE: Boxing Punching Bag}} 
    \end{subfigure}\\

    \begin{subfigure}{\textwidth}
        \centering
        \includegraphics[width=0.8\textwidth]{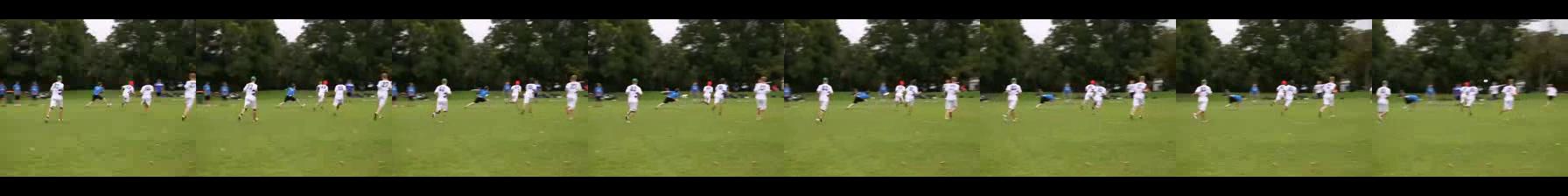}
        \caption*{GT: Frisbee Catch\\ \textcolor{green}{BIMM: Frisbee Catch}, \textcolor{red}{VideoMAE: Soccer Penalty}} 
    \end{subfigure}\\

    \begin{subfigure}{\textwidth}
        \centering
        \includegraphics[width=0.8\textwidth]{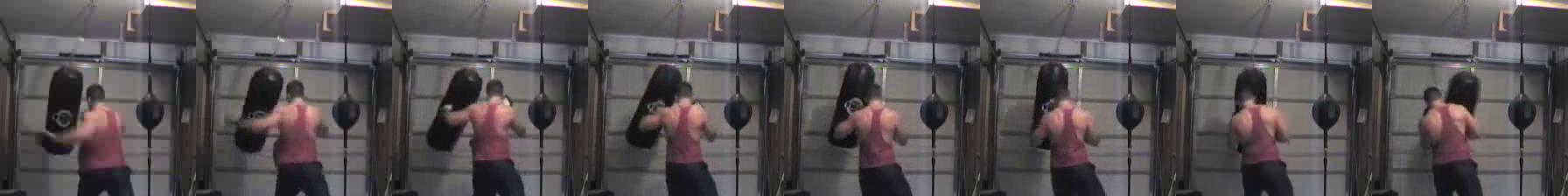}
        \caption*{GT: Boxing Punching Bag\\ \textcolor{red}{BIMM: Boxing Speed Bag}, \textcolor{red}{VideoMAE: Punch}} 
    \end{subfigure}
    
    \caption{Prediction examples of different models on UCF101. For each example drawn from the validation dataset, the predictions with \textcolor{green}{green} text indicating a correct prediction and \textcolor{red}{red} indicating the incorrect one. ``GT'' indicates the ground truth annotation of the video.} 
    \label{fig:quali}
\end{figure}

BIMM achieves significant performance improvements on various datasets compared to the baseline VideoMAE. To better understand how the model works, we select several examples from the UCF101 validation set, as shown in Figure. \ref{fig:quali}. 
VideoMAE makes incorrect judgments on all three examples. However, BIMM strengthens the learning of spatial information through ventral branch, and specifically learns the temporal representation in motion information through dorsal branch. Therefore, it can correctly identify the same action in different backgrounds (\eg, Punch vs. Boxing Punching Bag) or different actions in the same background (\eg, Frisbee Catch vs. Soccer Penalty).
It is worth noting that BIMM still makes mistakes in some difficult video examples, which include motions of small objects. This may be due to the high mask ratio in the masked modeling method.
We leave more detailed analysis and improvements for future work.

\subsection{Visualization of Video Reconstruction}

\begin{figure}[ht]
    \centering
    \begin{subfigure}{\textwidth}
        \centering
        \includegraphics[width=0.8\textwidth]{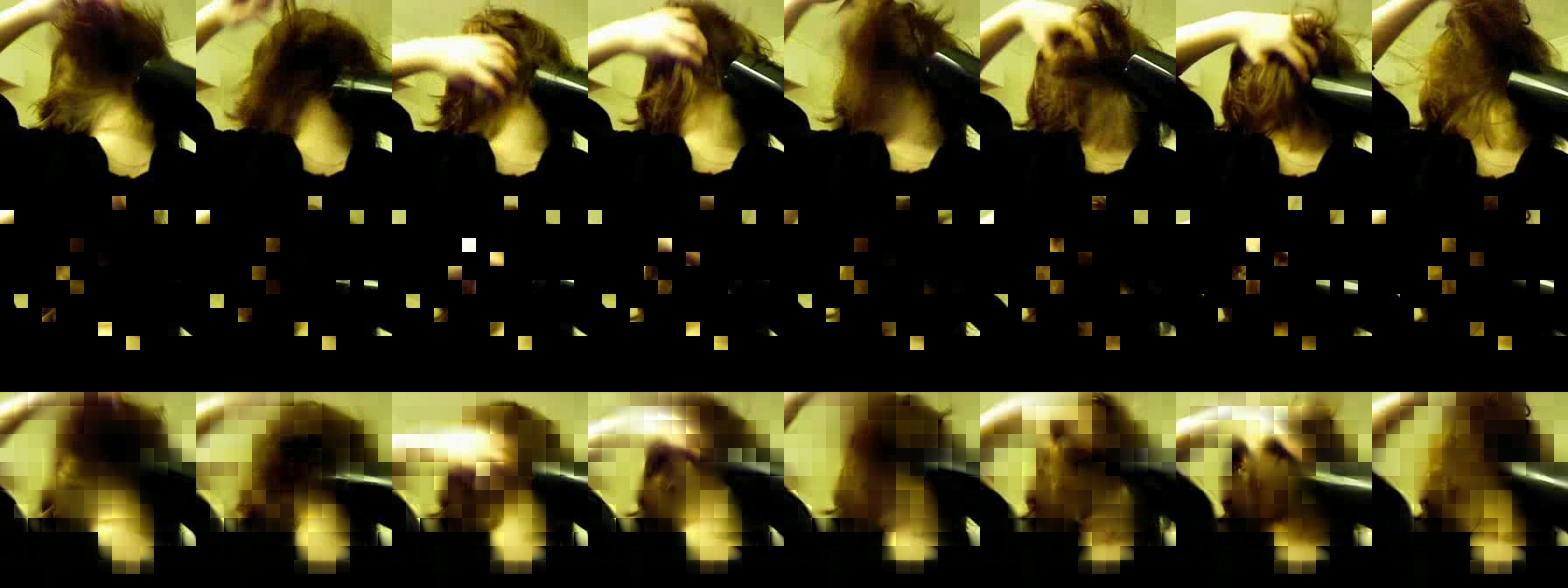}
        \caption*{(a) Blow Dry Hair} 
    \end{subfigure}\\ 

    \begin{subfigure}{\textwidth}
        \centering
        \includegraphics[width=0.8\textwidth]{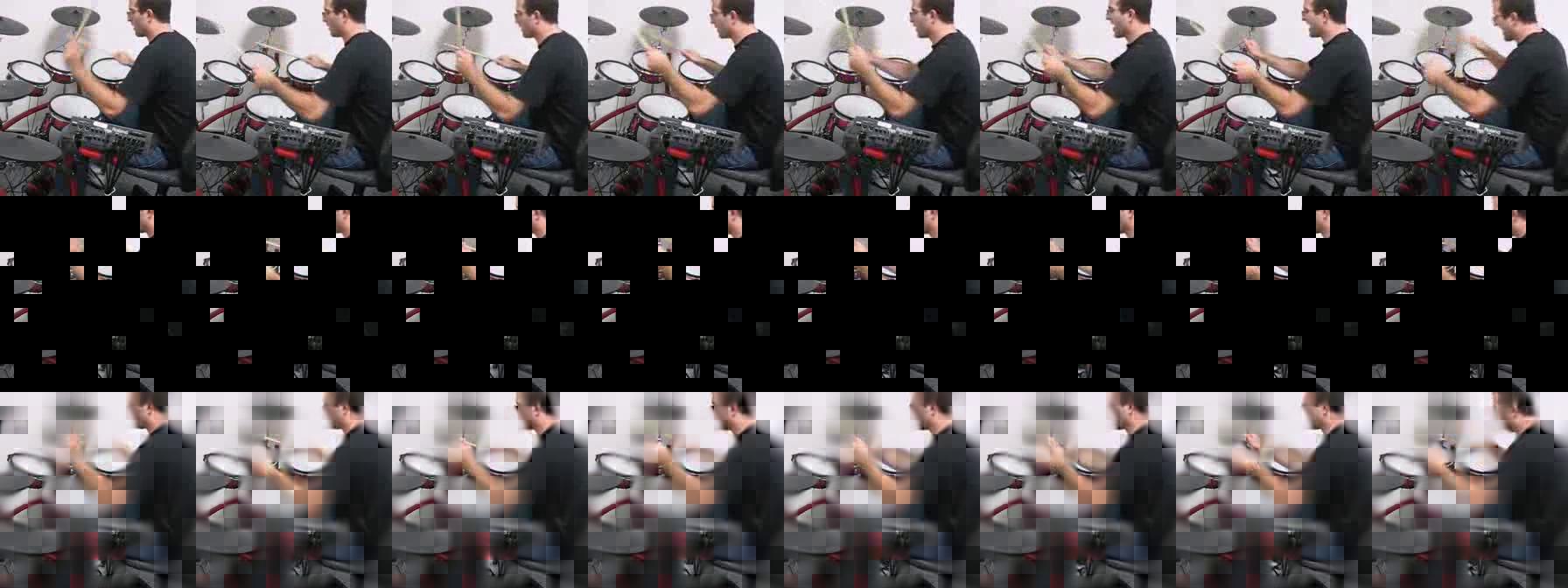}
        \caption*{(b) Drumming} 
    \end{subfigure}\\ 

    \begin{subfigure}{\textwidth}
        \centering
        \includegraphics[width=0.8\textwidth]{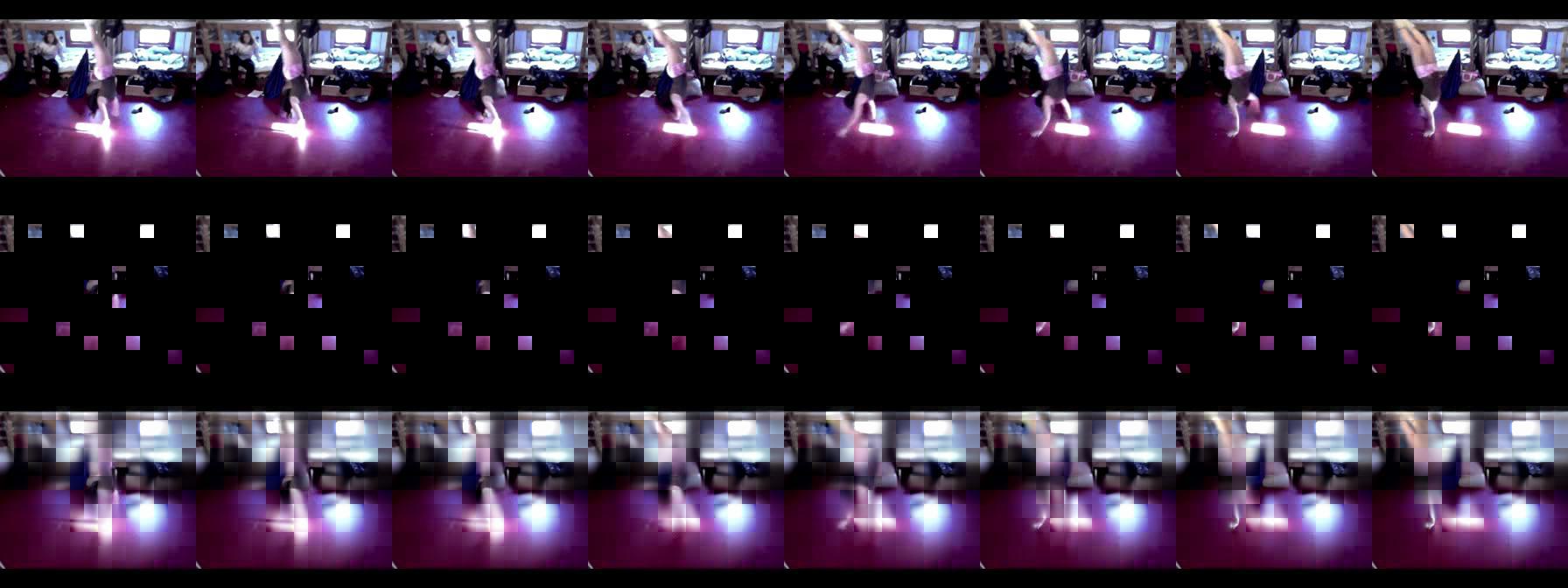}
        \caption*{(c) Handstand Walking} 
    \end{subfigure}

    \begin{subfigure}{\textwidth}
        \centering
        \includegraphics[width=0.8\textwidth]{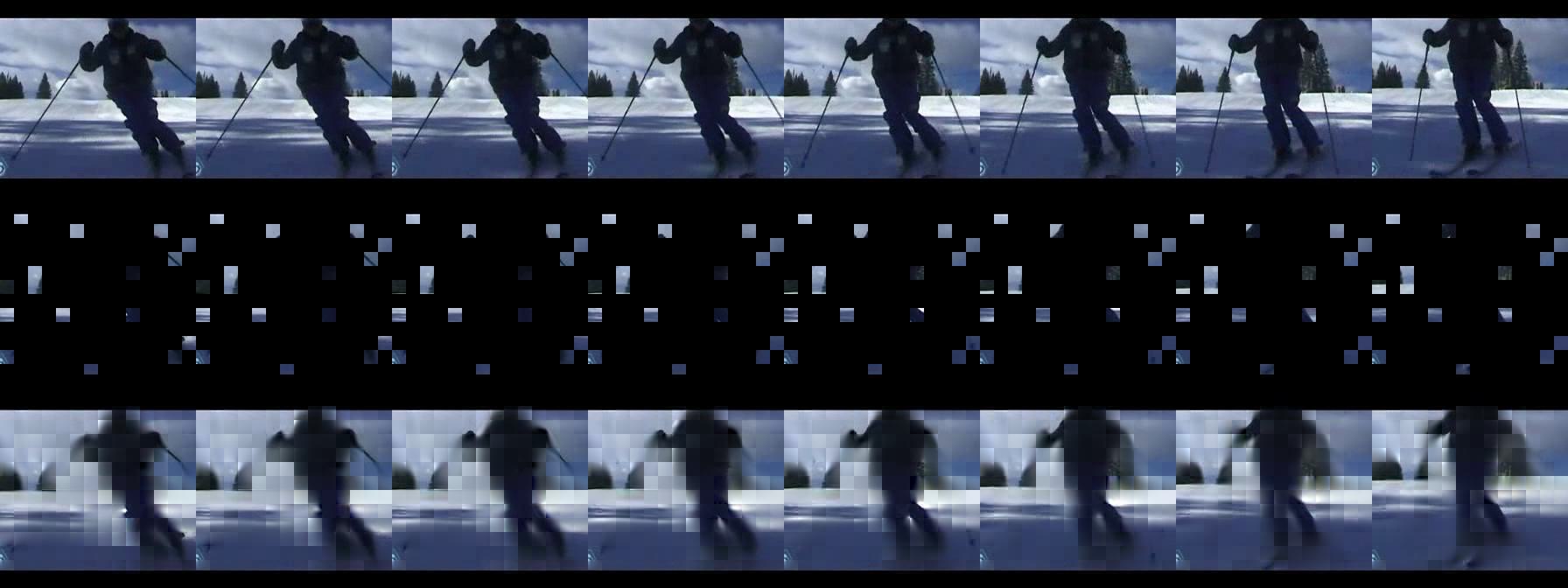}
        \caption*{(d) Skiing}
    \end{subfigure}
    \caption{Reconstruction results of videos on UCF101 validation set. We show the original video sequence, masked video sequence, and reconstructions of different videos. 
    Labels of each video are listed under each group of images.   Reconstruction of videos are predicted by the pretrained dorsal branch with a high masking ratio of 90\%, which indicates BIMM is able to learn comprehensive features even most patches are masked.} 
    \label{fig:vis_ucf}
\end{figure}

\begin{figure}[ht]
    \centering
    \begin{subfigure}{\textwidth}
        \centering
        \includegraphics[width=0.8\textwidth]{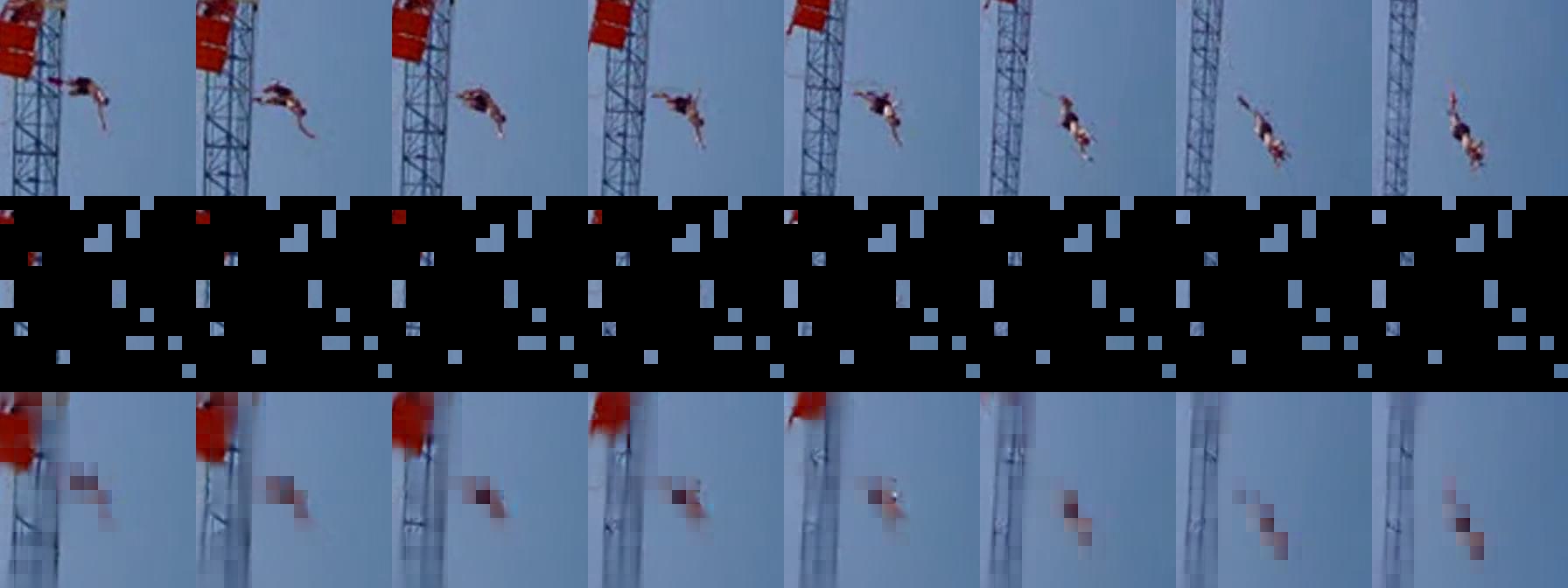}
        \caption*{(a) bungee jumping} 
    \end{subfigure}\\ 

    \begin{subfigure}{\textwidth}
        \centering
        \includegraphics[width=0.8\textwidth]{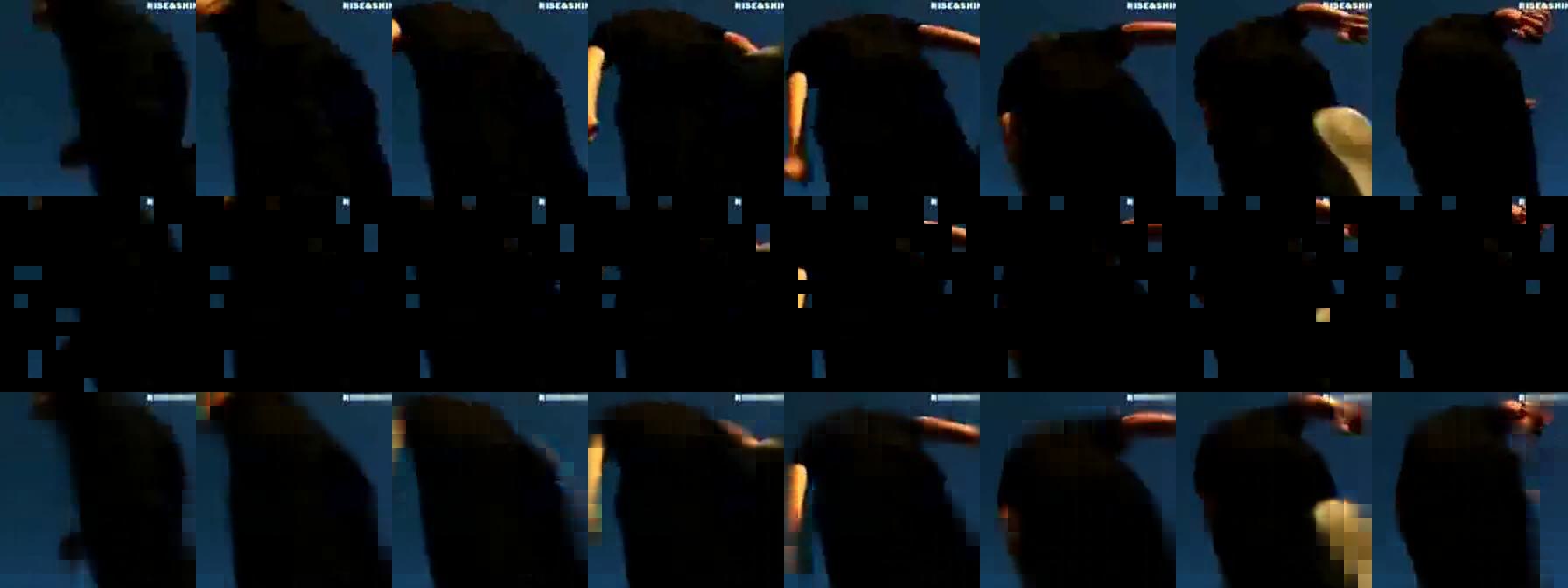}
        \caption*{(b) jump style dancing}
    \end{subfigure}\\ 

    \begin{subfigure}{\textwidth}
        \centering
        \includegraphics[width=0.8\textwidth]{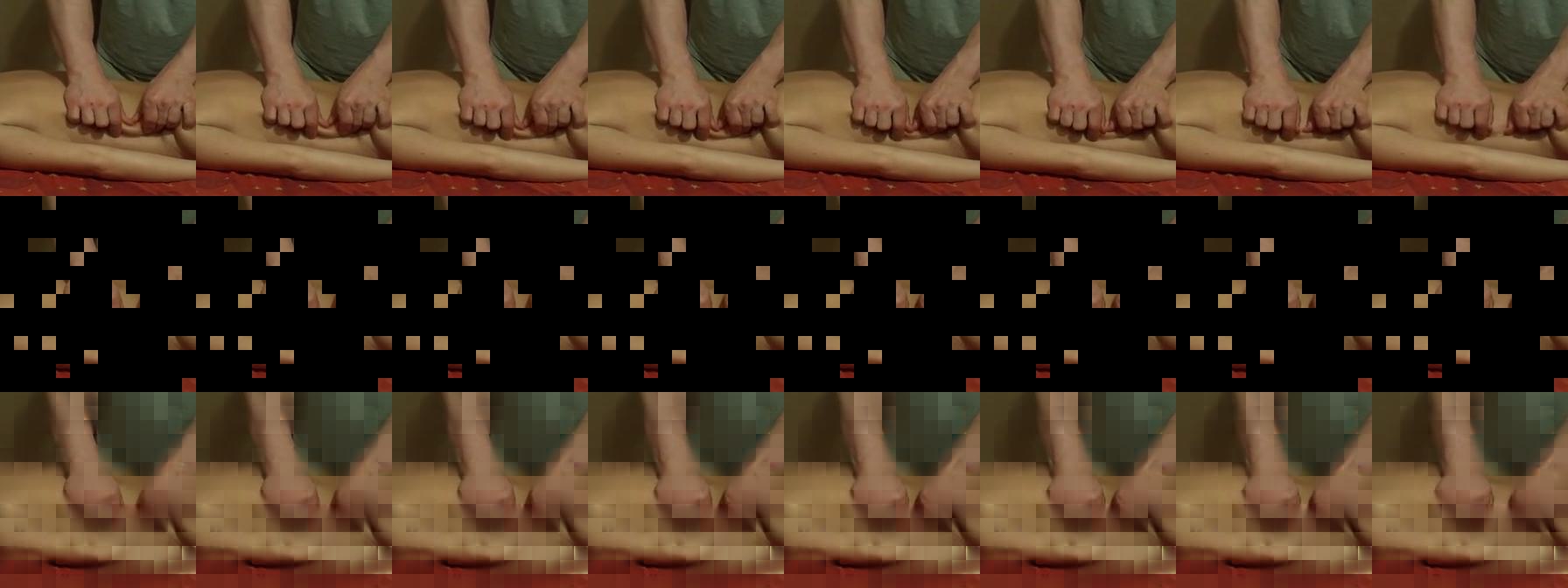}
        \caption*{(c) massaging back} 
    \end{subfigure}

    \begin{subfigure}{\textwidth}
        \centering
        \includegraphics[width=0.8\textwidth]{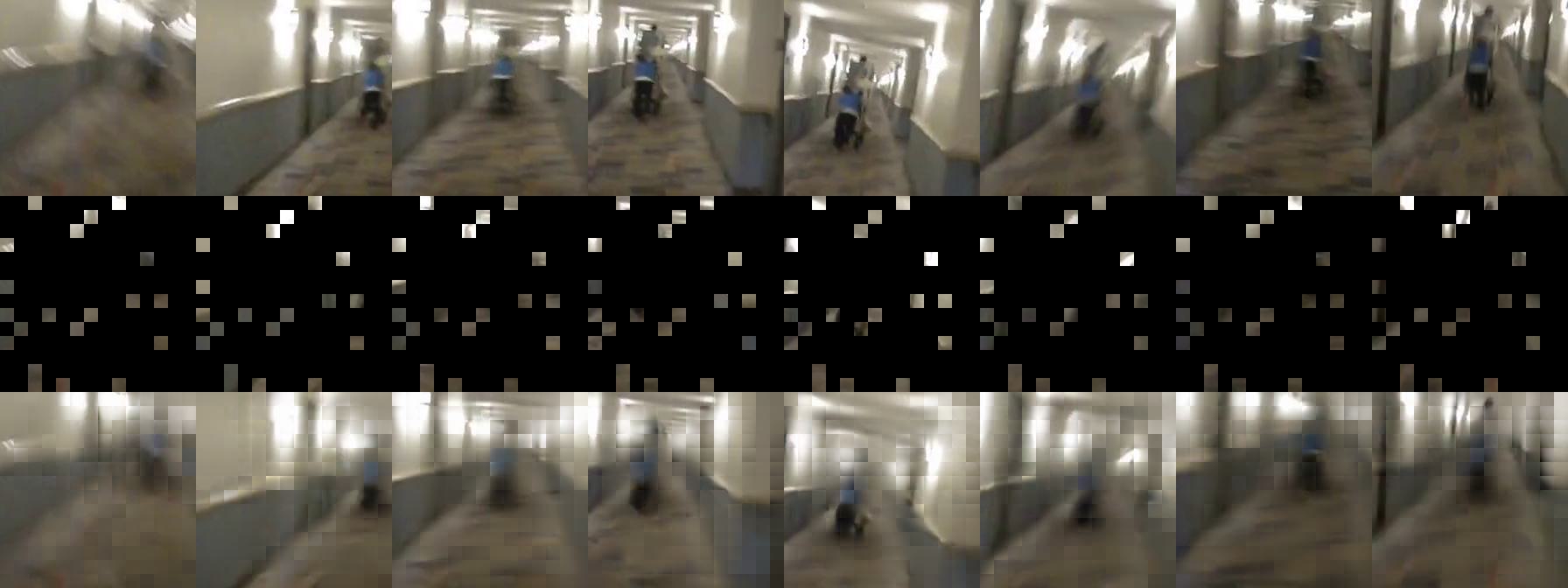}
        \caption*{(d) pushing cart} 
    \end{subfigure}
    \caption{Reconstruction results of videos on K400 validation set. We show the original video sequence, masked video sequence, and reconstructions of different videos. Labels of each video are listed under each group of images. Reconstruction of videos are predicted by the pretrained dorsal branch with a high masking ratio of 90\%, which indicates BIMM is able to learn comprehensive features even most patches are masked.} 
    \label{fig:vis_k400}
\end{figure}

We also show several examples of reconstruction in Figure. \ref{fig:vis_ucf} and \ref{fig:vis_k400}, with videos all randomly chosen from the UCF101 validation set and K400 validation set. Even under an extremely high masking ratio, BIMM can produce satisfying reconstructed results. 
Although it is not possible to reconstruct every precise details, the BIMM pretrained model is able to reconstruct human actions across a variety of scenarios, including half-body, full-body, human-object interactions, and even images with irregular compositions.
These examples imply that our BIMM is capable of learning more representative features that capture the holistic spatiotemporal structure in videos.

\end{document}